%% file: main.tex
\documentclass[letterpaper,twocolumn,10pt]{article}
\usepackage[table]{xcolor}
\usepackage{usenix-2020-09}

\input{preamble.tex}

\usepackage{amsmath}

\usepackage{filecontents}

\usepackage[]{hyperref}
\begin{document}

\date{}

\title{\name: A Systems Approach to Advancing Low-Bit LLM Quantization}


\author{
{\rm Yeonhong Park\textbf{*}\quad}
{\rm Jake Hyun$\textbf{*}$\quad}
{\rm Hojoon Kim\quad}
{\rm Jae W. Lee\quad} \vspace{5pt} \\
Seoul National University\quad
}

\maketitle
\def\thefootnote{\textbf{*}}\footnotetext{Equal contribution}\def\thefootnote{\arabic{footnote}}

\begin{abstract}
Quantization of Large Language Models (LLMs) has recently gained popularity, particularly for on-device settings with limited hardware resources. While efficient, quantization inevitably degrades model quality, especially in aggressive low-bit settings such as 3-bit and 4-bit precision. In this paper, we propose \name, an inference scheme that improves the quality of low-bit LLMs while preserving the key benefits of quantization: GPU memory savings and latency reduction. \name stores the residual matrix---the difference between full-precision and quantized weights---in CPU, and dynamically fetches the residuals for only a small portion of the weights. This portion corresponds to the salient channels, marked by activation outliers, with the fetched residuals helping to correct quantization errors in these channels. Salient channels are identified \emph{dynamically} at each decoding step by analyzing the input activations---this enables adaptation to the dynamic nature of activation distribution, thus maximizing the effectiveness of error compensation. We demonstrate the effectiveness of \name by augmenting state-of-the-art quantization methods. For example, \name reduces the perplexity of a 3-bit Llama-3-8B-Instruct model from 10.15 to 9.12—outperforming its 3.5-bit counterpart—while adding less than 0.0003\% to GPU memory usage and incurring only a 1.7\% inference slowdown on NVIDIA RTX 4050 Mobile.

\end{abstract}

\input{contents/1-intro}

\input{contents/2-background}

\input{contents/3-motivation}
\input{contents/4-system}

\input{contents/5-evaluation}
\input{contents/6-related-works}
\input{contents/7-conclusion}

\section*{Acknowledgments}
This work was supported by the National Research Foundation of Korea (NRF) grants funded by the Korea government (MSIT) (RS-2024-00340008, RS-2024-00405857). Jae W. Lee is the corresponding author.

\bibliographystyle{plain}
\bibliography{references.bib}

\end{document}

%% file: preamble.tex
\usepackage{multirow}
\usepackage{multicol}
\usepackage{tabu}
\usepackage{graphicx}
\usepackage{makecell}
\usepackage{amsthm}
\usepackage{mathtools}
\usepackage{microtype}
\usepackage[export]{adjustbox}
\usepackage[T1]{fontenc}
\usepackage[scaled=0.825]{beramono}
\usepackage{hhline}
\usepackage[ruled,vlined,linesnumbered,algo2e]{algorithm2e}
\usepackage{xspace}
\usepackage{lipsum}
\usepackage{comment}
\usepackage[super]{nth}
\usepackage{array}
\usepackage{balance}

\usepackage{amsmath,amsfonts}
\usepackage{algorithmic}
\usepackage{textcomp}
\usepackage{listings}
\usepackage{tikz}
\usepackage{comment}
\usepackage[bb=boondox]{mathalfa}
\usepackage[figuresright]{rotating}
\usepackage{bm}
\usepackage{graphicx} 
\usepackage{makecell}
\usepackage{nicefrac}

\newcommand{\para}[1]{\noindent \textbf{#1.}}

\newcommand*\circled[1]{\tikz[baseline=(char.base)]{            \node[shape=circle,fill,inner sep=1pt] (char) {\textcolor{white}{#1}};}}
\DeclareMathOperator*{\argmin}{argmin}   

\newcommand{\rev}[1]{\textcolor{black}{#1}}

\newcommand{\name}{DecDEC\xspace}
\newcommand{\rbw}{\textbf{$R_{bw}$}\xspace}

%% file: contents/1-intro.tex
\section{Introduction}
\label{sec:introduction}
Recent advancements in Large Language Models (LLMs) based on the Transformer architecture~\cite{attention} have shown great potential to reshape our daily lives~\cite{fewshot, llama, gpt-4, phi-3, gemini}. 
However, their deployment costs pose a significant challenge, as large model sizes increase memory requirements and latency, limiting their use cases~\cite{efficient}.
Quantization is a promising solution for reducing the LLM deployment costs~\cite{survey, llama-3-quant}. By lowering the model precision, quantization addresses both memory limitations and inference latency. The importance of quantization is pronounced for on-device deployments, where strict memory budgets often make model compression mandatory rather than optional.
These scenarios typically require careful tuning of quantization levels to achieve an optimal balance between model size and quality~\cite{layerwise-1, layerwise-2, channelwise-1, zeroq, exllamav2}.

While quantization may allow gigantic LLMs to fit into small-memory devices, it often leads to model quality degradation due to the inevitable loss of information. This is especially true for low-bit settings, such as 3-bit and 4-bit quantization, which are often used to accommodate the parameter sizes of LLMs~\cite{sqllm, awq, gptq}. This raises a key research question that this paper addresses: given a quantized LLM configured with the best possible effort under the memory budget, is there a way to recover the quality loss caused by quantization?

Leveraging external memory offers a potential solution to this problem. Specifically, on heterogeneous computing platforms where the CPU and GPU are connected via a PCIe interconnect---a common architecture in desktops and laptops---CPU memory becomes a viable option. Additional information that may be used to mitigate quantization errors can be stored in CPU memory and fetched at runtime, avoiding any additional GPU memory overhead.

However, utilizing CPU memory for GPU inference presents a critical challenge: the slow data transfer between the CPU and GPU can create a bottleneck for inference latency. To mitigate this issue, the volume of data transferred must be carefully controlled. Therefore, designing a system that effectively utilizes CPU memory requires the identification of the minimal set of additional information that can substantially enhance the quality of quantized models, while keeping the impact on latency minimal.

A clue for this problem comes from the well-known fact that not all channels in the weight matrix are equally important for quantization.
Some channels, referred to as \emph{salient channels}, are more critical, primarily due to the presence of activation outliers~\cite{llmint8, quantizable, rethinking}. When certain input activation values are particularly large, the quantization errors in the corresponding weight channels—those multiplied by these large activation values—are amplified, making the channels salient. By identifying these channels and selectively fetching error compensation terms from CPU memory, we can maximize the quality boost while minimizing data transfer.

Hence, the identification of the salient channels is essential for overcoming the bandwidth limitation, allowing for the transfer of only impactful information. Previous works that attempt to improve quantization algorithms by addressing activation outliers analyze the activation value distribution on a calibration data to predetermine the salient channels~\cite{llmint8, owq, slim-llm, awq}. This approach is suboptimal as it statically designates certain channels as salient throughout the inference, while the real distribution of activation values---and thus the distribution of salient channels---changes dynamically at each decoding step. Accounting for this dynamic nature is crucial for the accurate identification of salient channels.

Thus, in this paper, we introduce \name (\textbf{Dec}oding with \textbf{D}ynamic \textbf{E}rror \textbf{C}ompensation), an inference scheme for quantized LLMs that dynamically identifies salient channels and compensates for quantization errors in these channels, in real time. \name enhances model quality while fully preserving the two main benefits of quantization: GPU memory savings and latency reduction. To achieve this, \name stores the residuals of the quantized weight matrices in CPU memory, fetching only the parts that correspond to the dynamically identified salient channels for error compensation. Although small in size, these residuals provide a significant quality boost. This dynamic error compensation is performed concurrently with inference by an optimized GPU kernel, ensuring that all additional operations are seamlessly integrated into the existing workflow, minimizing inference slowdown. Below are the key contributions of our work:

\begin{itemize}
    \item We provide an in-depth analysis on the dynamic nature of activation outliers in LLM inference.
    \item We present \name, an inference scheme that enhances quantized LLMs by dynamically identifying salient channels and compensating quantization errors in them.
    \item We introduce a tuner for \name that recommends system parameters to satisfy a target latency bound.
    \item We evaluate \name across five different consumer-grade GPUs, demonstrating significant quality improvement with minimal memory and latency overhead.
\end{itemize}

%% file: contents/2-background.tex
\section{Background}
\label{sec:background}

\begin{figure}[t]
    \centering
    \includegraphics[width=\columnwidth]{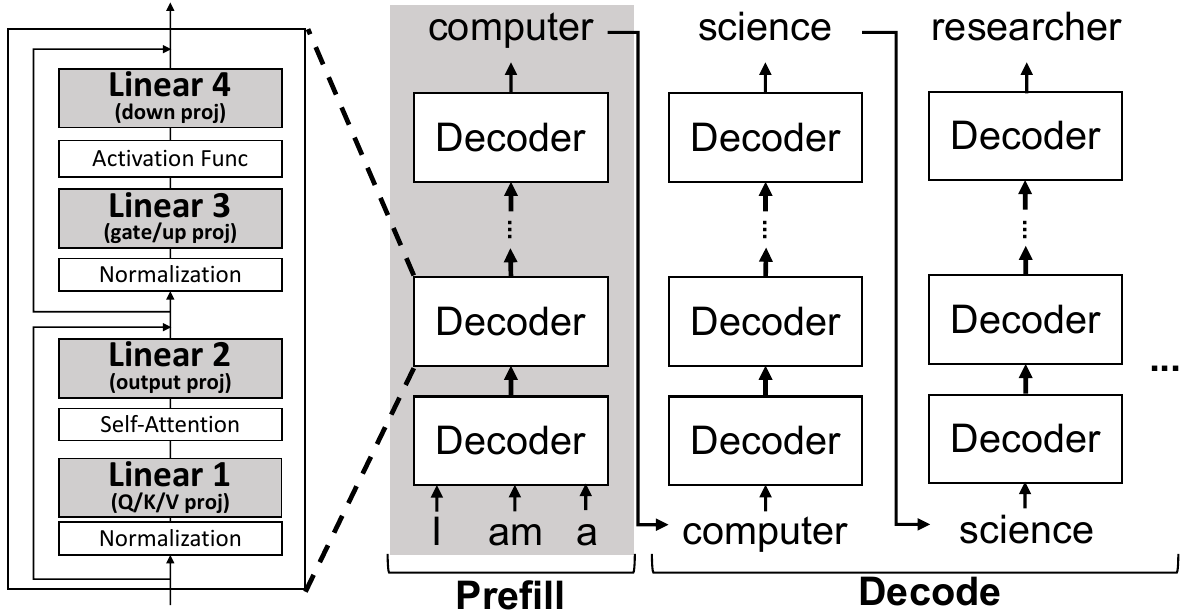}
    \caption{LLM inference.}
    \label{fig:llm}
\end{figure}

\subsection{LLM Inference}
\label{sec:llm-inference}
Figure~\ref{fig:llm} presents a description of the modern LLM architecture and its inference flow. LLMs consist of multiple Transformer decoder blocks~\cite{attention}, each containing multiple linear layers, which account for most of the inference time, alongside other components like self-attention and normalization.

LLM inference involves two phases: prefill and decode. In the prefill phase, all input tokens (e.g., 'I', 'am', 'a') are processed in parallel to generate a single output token (e.g., 'computer'). The decode phase begins subsequently, where the output token of the previous step is fed back into the model to generate the next token, repeated until the end of the sequence. This sequential nature of the decode phase makes it the primary latency bottleneck.

The decode phase is particularly memory-bound, as only one token is processed at a time, reducing the linear layers to GEMV operations. In data center settings, this issue can be alleviated by batching together multiple queries~\cite{orca, vllm}. However, this usually cannot be applied to on-device inference, where LLMs serve only individual users.

\subsection{LLM Quantization}
\label{sec:llm-quantization}
Quantization for LLMs---a popular compression technique that reduces both memory usage and inference latency---can be categorized into two main types~\cite{survey, llama-3-quant}: weight-activation quantization~\cite{zeroquant, llmint8, omniquant, smoothquant, llmqat, emnlp} and weight-only quantization~\cite{gptq, awq, sqllm, spqr, owq, quip, quip-sharp, aqlm, qlora, peqa}. Each are suited to different inference scenarios. Weight-activation quantization is primarily used in datacenter settings, where both memory and computational costs must be minimized to improve throughput. Quantizing both weights and activations allows for the efficient use of low-precision arithmetic units (e.g., INT4, INT8, FP8) available on modern GPUs~\cite{h100_datasheet}. In contrast, for on-device inference, weight-only quantization is the preferred approach~\cite{survey, sqllm}. In this apporach, quantized weights are loaded from memory and dequantized on-the-fly to full precision (i.e., FP16), before being multiplied with the full-precision activations~\cite{lutgemm}. Though it only reduces memory traffic, this is sufficient to speed up on-device inference where memory is the bottleneck, as discussed in Section~\ref{sec:llm-inference}.

Weight-only quantization can be further divided into two sub-categories: quantization-aware training (QAT)~\cite{llmqat, qlora,peqa}, and post-training quantization (PTQ)~\cite{gptq, awq, sqllm, spqr, owq, quip, quip-sharp, aqlm}. While QAT yields better results by retraining to reduce quantization errors, its cost makes it impractical for many end-users~\cite{sqllm, survey}. As a result, PTQ---requiring no retraining---has become the preferred method for on-device LLM inference. Therefore, in this paper, we specifically focus on weight-only PTQ for on-device inference.

%% file: contents/3-motivation.tex
\section{Augmenting Quantized LLM with CPU Memory}
\label{sec:strategy}

In this section, we propose leveraging CPU memory---an often underutilized resource in LLM inference, as GPUs are the de facto standard processors for this workload---as a means to augment quantized LLMs. Section~\ref{sec:concept} introduces the concept of CPU-augmented quantized LLMs while Section~\ref{sec:opportunity} and Section~\ref{sec:challenge} explore its opportunities and challenges.

\begin{figure}[t]
    \centering
    \includegraphics[width=\columnwidth]{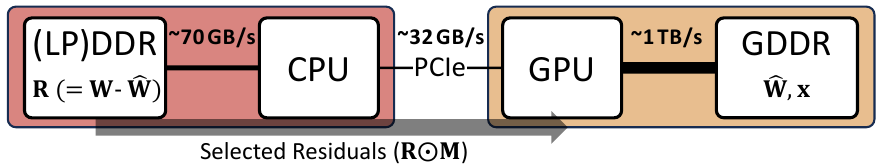}
    \caption{CPU-augmented inference for quantized LLMs.}
    \label{fig:concept}
\end{figure}

\subsection{Concept}
\label{sec:concept}
\para{Goal}
We aim to leverage CPU memory to improve quantized LLM quality without additional GPU memory costs. Quantization trades model quality for a smaller memory footprint. In constrained settings, practitioners optimize this trade-off within a fixed GPU memory budget by selecting a uniform bitwidth or applying fine-grained strategies such as layer-wise~\cite{layerwise-1,layerwise-2,zeroq} or channel-wise~\cite{channelwise-1, exllamav2} allocation. Our goal is to then further improve quality post hoc by utilizing CPU memory, without increasing GPU memory usage.

\para{Basic Mechanism}
Figure~\ref{fig:concept} illustrates our concept of leveraging CPU memory to enhance quantized LLMs. We primarily target desktop or laptop platforms, where the GPU is connected to the CPU via a PCIe interconnect. As in conventional inference systems, the quantized weight parameters ($\widehat{\mathbf{W}}$) and activations ($\mathbf{x}$) are kept in GPU memory. The difference here is that $\mathbf{R}$---the residual between the original full-precision weights and the quantized weights---is stored in CPU memory. During the decode phase, residuals are fetched from the CPU to help compensate for quantization errors, potentially improving model quality. Due to the limited bandwidth of PCIe, which is typically an order of magnitude lower than GPU memory bandwidth (e.g., 32 GB/s vs. 1 TB/s), fetching the entire residual matrix would incur a prohibitive latency bottleneck. Therefore, only a small subset of residuals should be fetched in a selective manner. In short, this process augments each linear operation of quantized LLM from $\widehat{\mathbf{W}}\mathbf{x}$ to $(\widehat{\mathbf{W}}+\mathbf{R\odot M})\mathbf{x}$, where $\mathbf{M}$ is a binary mask that sparsifies $\mathbf{R}$.
 
\para{Key Research Question}
A key research question in designing an effective CPU-augmented inference system for quantized LLMs is determining a subset of residuals, or mask $\mathbf{M}$. A good mask $\mathbf{M}$ should: 1) select portions of the residuals that contribute most to improving model quality within the bandwidth constraints, and 2) maintain a structured form that minimizes indexing overhead and ensures effective residual transfer, while enabling efficient processing on the GPU.

\begin{figure}[t]
    \centering
    \includegraphics[width=1.0\linewidth]{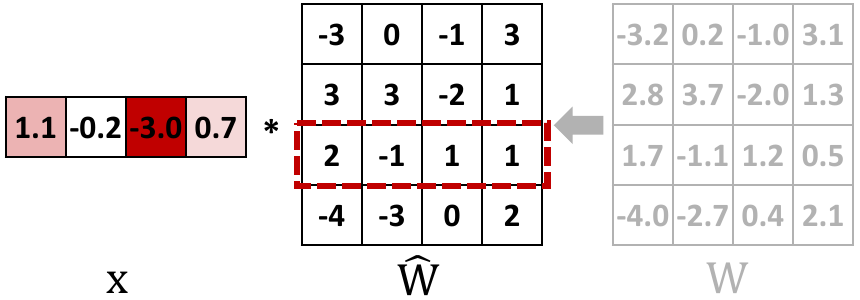}
    \caption{Activation outlier issue in weight quantization.}
    \label{fig:activation_outliers}
\end{figure}

\subsection{Opportunity: Not All Residuals Are Equally Important}
\label{sec:opportunity}

Some residuals are more important than others---an opportunity that can be leveraged to determine $\mathbf{M}$. This opportunity arises from the presence of activation outliers (i.e., activation values with large magnitudes), a well-known phenomenon in LLM inference~\cite{llmint8, owq, awq, slim-llm, quantizable, rethinking}. When certain activation values are noticeably large, even small quantization errors in the corresponding weight channels can be multiplied and amplified, leading to considerable perturbations in the output. Figure~\ref{fig:activation_outliers} illustrates this issue. In this example, the third input channel (third row) of the weight matrix is multiplied by the activation outlier, -3.0. We refer to such channels as \emph{salient} channels. Constructing $\mathbf{M}$ at the input channel granularity based on the magnitude of input activations can satisfy two key conditions for an effective mask: selecting impactful portions of the residuals and maintaining a structured form.

\begin{figure}[t]
    \centering
    \includegraphics[width=\linewidth]{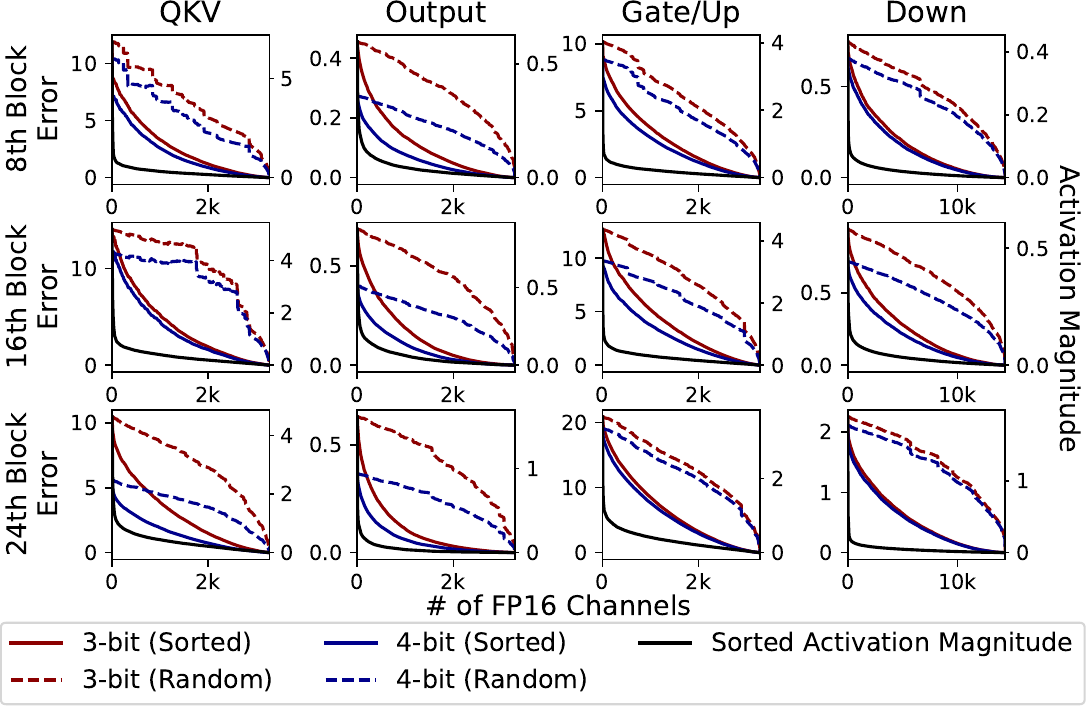}
    \caption{Quantization error reduction trends observed when replacing the input channels of quantized weights with FP16 values sequentially in sorted order (solid lines) and random order (dotted lines) for the 8th, 16th and 24th decoder block of Llama-3-8B-Instruct. The distribution of activation magnitudes in sorted order is also shown (black lines).}
    \label{fig:cumul_errors}
\end{figure}

Indeed, compensating for errors in salient channels using the corresponding residuals is highly effective. Figure~\ref{fig:cumul_errors} illustrates how the quantization error, defined as the mean squared error between the computation result with FP16 weights (${\mathbf{W}}\mathbf{x}$) and quantized weights ($\widehat{\mathbf{W}}\mathbf{x}$), is reduced by sequentially replacing the input channels of the quantized weight with their corresponding FP16 values. For this analysis, we evaluate 3-bit and 4-bit versions of the LLaMA-3-8B-Instruct model~\cite{llama-3}, quantized using the state-of-the-art method AWQ~\cite{awq}, with a text sample from the C4 dataset~\cite{c4} as the input prompt. All four linear layers in the 8th, 16th, and 24th decoder blocks are included in this evaluation. The quantization errors for both the 3-bit and 4-bit models drop rapidly when we progressively compensate for channels in descending order of their activation magnitudes (solid red and solid blue lines). This trend closely follows the activation magnitude distribution (solid black line), which represents the sorted activation magnitudes in descending order. In contrast, the reduction in quantization error is significantly slower when input channels are compensated in random order, as shown by the dotted lines. This highlights the importance of prioritizing salient channels based on activation magnitude.

\begin{figure}[t]
    \centering
    \includegraphics[width=\linewidth]{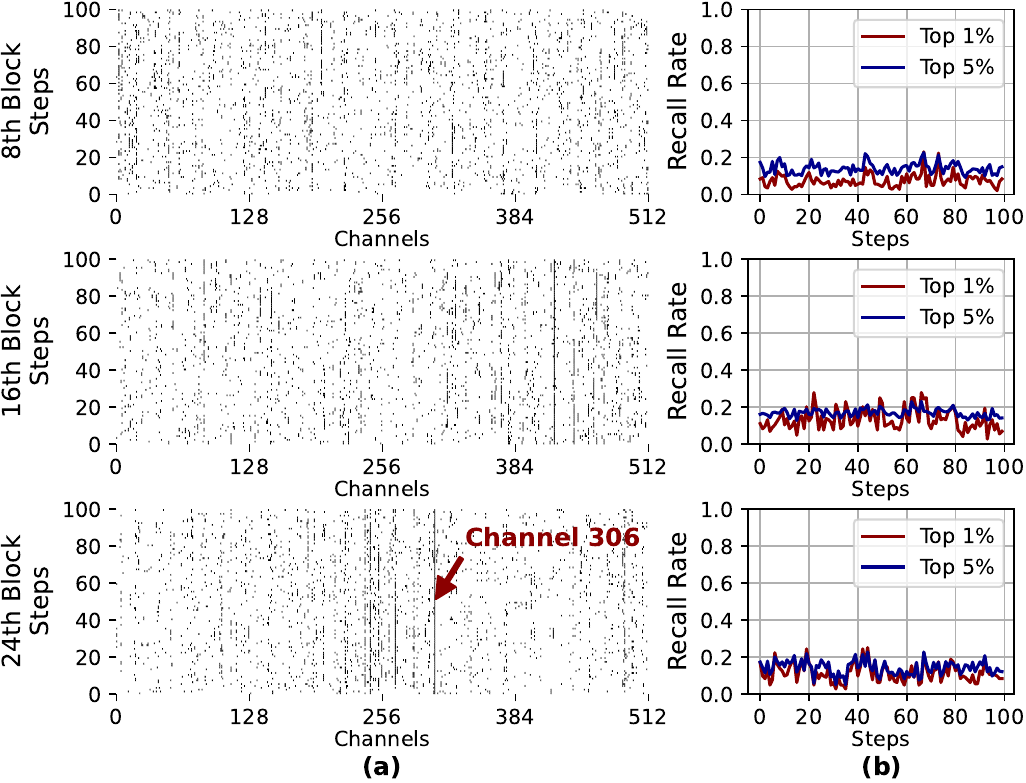}
    \caption{\rev{(a) Distribution of activation outliers (top 5\%) and (b) Recall rate of static analysis-based outlier identification for the true top 1\% and 5\% outliers, across 100 decoding steps. The down projection layers in the 8\textsuperscript{th}, 16\textsuperscript{th}, and 24\textsuperscript{th} decoder blocks of Llama-3-8B-Instruct model are used for profiling.}}
    \label{fig:dynamic_activation_outliers}
\end{figure}

\subsection{Challenge: Dynamic Nature of Activation Outliers}
\label{sec:challenge}
\rev{Identifying salient channels is challenging because the distribution of activation outliers changes dynamically by nature. While it is possible to infer salient channels by statically analyzing activation value statistics on a small calibration set~\cite{llmint8, owq, slim-llm, awq}, such static approaches are suboptimal. Figure~\ref{fig:dynamic_activation_outliers}(a) shows the distribution of activation outliers---defined as activations with the top 5\% magnitudes---in the down projection layer of the 8\textsuperscript{th}, 16\textsuperscript{th}, and 24\textsuperscript{th} decoder blocks of the Llama-3-8B-Instruct model over 100 decoding steps, using a text sample for C4 dataset as the input prompt~\cite{c4}. For visibility, only the first 512 channels are shown. While some channels (e.g., Channel 306 in 24th block, highlighted by an arrow) consistently exhibit high activation magnitudes and remain persistent outliers, the outlier distribution generally shows significant irregularity across decoding steps.}

To quantify the dynamic nature of activation outliers, we calculate the recall rate of the top 1\% and top 5\% outliers identified through static analysis using a calibration set, compared to the true top 1\% and top 5\% outliers (ground truth)  observed at each decoding step. We use a subset of the Pile dataset~\cite{pile} for calibration, following prior work~\cite{awq}. Specifically, we profile the average of the mean square of each activation value and use this as a metric for identifying outliers. Figure~\ref{fig:dynamic_activation_outliers}(b) presents the results, showing that the recall rate remains low ($\sim$20\%) for both the top 1\% and top 5\% outliers. This highlights a clear limitation of static analysis, as it misses the majority of outliers at runtime, emphasizing the need for dynamic identification of salient channels.

%% file: contents/4-system.tex
\section{\name Design}
\label{sec:design}

\begin{figure*}[t]
    \centering
\includegraphics[width=\textwidth]{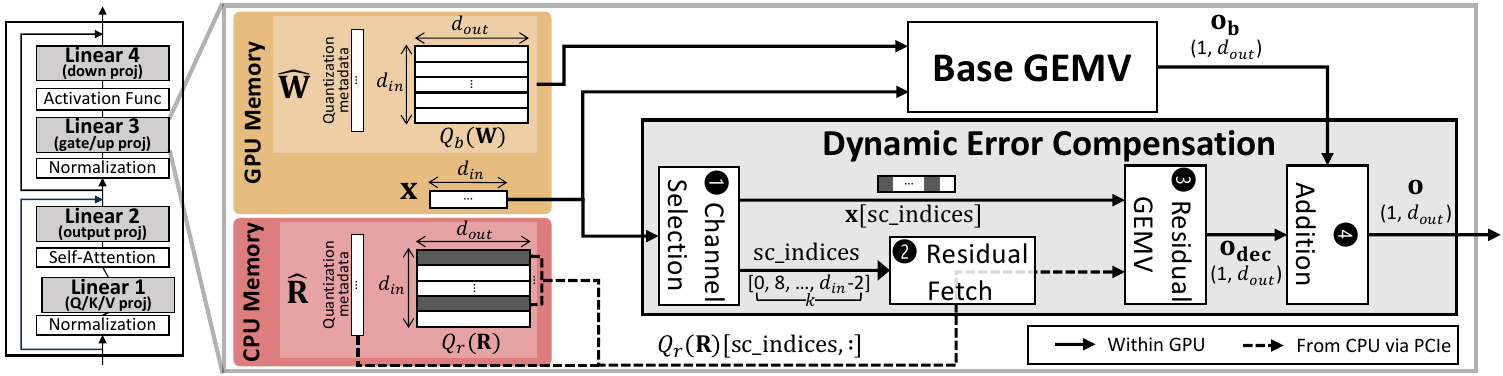}
    \caption{\name overview.}
    \label{fig:overview}
\end{figure*}

\subsection{Overview}

In this section, building on the opportunity outlined in Section~\ref{sec:opportunity} and addressing the challenge described in Section~\ref{sec:challenge}, we propose \name, a CPU-augmented inference system for quantized LLMs that performs decoding with dynamic error compensation. Figure~\ref{fig:overview} presents an overview of \name. During the decode phase, \name augments each linear layer, essentially a GEMV operation, with dynamic error compensation.
To produce the final output $\mathbf{o}$, \name adds an error compensation term $\mathbf{o_{\text{dec}}}$ to the base GEMV result $\mathbf{o_{\text{b}}=\widehat{W}x}$.
$\mathbf{o_{dec}}$ is computed by multiplying the input vector with a subset of weight residuals selectively fetched from the CPU. This selection is performed \emph{dynamically}, fully accounting for the variability of input vectors. To maximize the number of residual values fetched under PCIe bandwidth constraints, \name stores and retrieves a quantized version of the residuals ($\mathbf{\widehat{R}}$), comprising the quantized values $Q_{r}(\mathbf{R})$ and associated metadata, instead of the full-precision residuals ($\mathbf{R}$). Here, $Q_r$ is a quantizer that maps full precision residuals to low-bit form and differs from the base quantizer used for the weights, $Q_b$.

The dynamic error compensation process consists of four sequential steps. \circled{1} First, by investigating the input activation vector, \name creates \texttt{sc\_indices}, a list of salient channel indices. The number of salient channels to compensate, $k$, is a preconfigured parameter. This step is essentially a Top-K operation that selects the values in the input activation vector with the largest magnitudes. \circled{2} Next, a portion of the quantized residuals corresponding to the salient channels, $Q_{r}(\mathbf{R})[\texttt{sc\_indices}, :]$ (along with the necessary quantization metadata), is fetched from the CPU via PCIe. \circled{3} The fetched residuals are then multiplied by the sparsified activation vector ($\mathbf{x}[\texttt{sc\_indices}]$), producing $\mathbf{o_{dec}}$. \circled{4} Finally, the resultant $\mathbf{o_{dec}}$ is added to the base GEMV result $\mathbf{o_{b}}$, producing the final output, $\mathbf{o}$. All steps run in parallel with the base GEMV on a different GPU stream, and must be highly efficient to remain hidden within base GEMV runtime.


The following sections provide details of each component of \name. Section~\ref{sec:residual_quantization} explains how \name performs the residual quantization. Section~\ref{sec:gpu_implementation} details the GPU implementation of the dynamic error compensation. Section~\ref{sec:tuner} illustrates how \name configures system parameters, including $k$, the number of channels to compensate.

\begin{figure}[t]
    \centering
    \includegraphics[width=0.95\columnwidth]{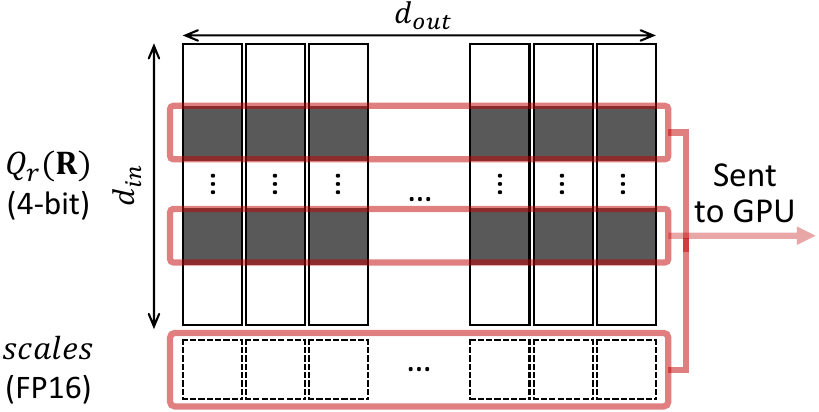}
    \caption{Quantization of weight residual.}
    \label{fig:quantization}
\end{figure}

\subsection{Residual Quantization}
\label{sec:residual_quantization}
Figure~\ref{fig:quantization} depicts the quantization scheme for the residuals. \name employs 4-bit quantization for each output channel (i.e., column) of the residuals. To minimize metadata, symmetric uniform quantization is used. This approach requires only a single scalar scale factor as metadata for each output channel. The residual quantizer for the $i$-th output channel ($Q_{r,i}$) is defined as:
\begin{equation*}
Q_{r,i}(r) = \text{clip}\left(\text{round\_to\_int}\left(\frac{r}{S_i}\right), -7, 7\right).
\end{equation*}
$S_{i}$, the scale factor, is determined through a grid search as the value that minimizes the mean squared error between the original and quantized weights.


Using this quantizer, each floating-point residual value $r$ is projected to an integer between -7 and 7. At runtime, the selected input channels of the quantized residuals (highlighted in Figure~\ref{fig:quantization}) and all the scale factors are fetched from the CPU. Each input channel of the quantized residuals, as well as the scales, are stored contiguously in CPU memory, enabling coalesced data transfers.

\subsection{Efficient Implementation of Dynamic \\ Error Compensation}
\label{sec:gpu_implementation}
The top priority in implementing dynamic error compensation is ensuring low latency, allowing its execution to remain hidden within the base GEMV execution time. To fetch a sufficient number of residual channels within this short time window, \name introduces three key software optimization strategies: 1) zero-copy residual fetching, 2) fast approximate Top-K for channel selection, and 3) kernel fusion.

\begin{figure*}[t]
    \centering
    \includegraphics[width=\textwidth]{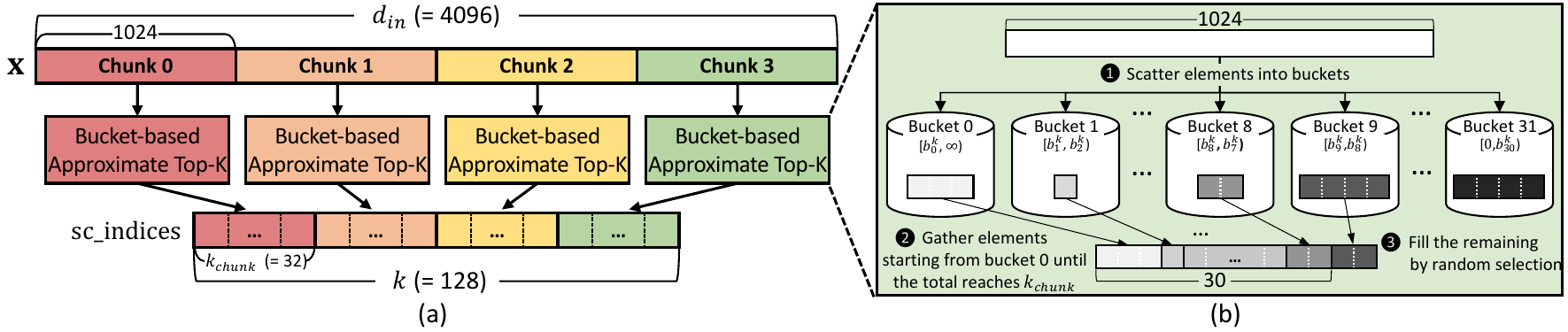}
    \caption{Fast approximate Top-K operation of \name.}
    \label{fig:topk}
\end{figure*}

\begin{figure}[t]
    \centering
    \includegraphics[width=\columnwidth]{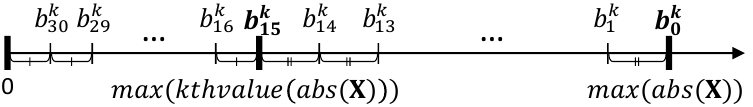}
    \caption{\name’s approximate Top-K bucket boundaries.}
    \label{fig:boundary}
\end{figure}

\para{Zero-Copy Residual Fetch}
\name leverages CUDA zero-copy~\cite{zero-copy}, instead of commonly used API functions such as \texttt{cudaMemcpy()} or \texttt{cudaMemcpyAsync()}, to fetch the residuals from CPU. These APIs rely on the direct memory access (DMA) engine for data transmission, which is efficient at transferring large data blocks but suboptimal for smaller transfers due to the DMA setup overheads. Fetching residuals, however, falls into the latter category. The granularity of residual fetching occurs at the row level of the quantized residual matrix. With 4-bit quantization and typical row lengths of a few thousand to tens of thousands, each data block transfer is only a few tens of KBs. For optimal PCIe bandwidth utilization, the data block size should ideally be at least a few hundred KBs (e.g., 256 KB)~\cite{pytorch_direct, pearson}. In zero-copy access, the GPU directly sends cacheline-sized memory requests, making it suitable for fine-grained data access. While zero-copy access has the disadvantage of occupying GPU cores to generate memory requests—potentially slowing down other concurrently running kernels—this is not a major issue for \name. The concurrent kernel for our case, the base GEMV, is typically memory-bound, so using fewer cores for this kernel is unlikely to have significant impact on its execution time.

\para{Fast Approximate Top-K for Channel Selection}
For channel selection, instead of an exact Top-K operation, \name employs an approximate Top-K method that is fast and GPU-friendly while maintaining precision. Figure~\ref{fig:topk} illustrates the approach using an example where 128 elements are selected from a 4096-dimensional activation vector ($d_{in} = 4096$, $k = 128$). As shown in Figure~\ref{fig:topk}(a), instead of a single global selection, \name partitions the input into four contiguous 1024-dimensional chunks and performs a local Top-$k_{chunk}$ selection within each chunk. In this example, $k_{chunk}=k/4=32$. The locally selected elements are then concatenated to form the final result. Although this introduces approximation, it significantly reduces latency by avoiding global synchronization---each local selection is handled independently by a thread block. A larger chunk size can improve efficiency but may increase the approximation error; we set the chunk size to 1024 to balance this trade-off effectively.

For each local selection, \name uses a variant of the bucket-based Top-K algorithm~\cite{bucket}. Figure~\ref{fig:topk}(b) illustrates this process. \circled{1} First, the 1024 elements in a chunk are scattered into buckets based on their magnitudes, with bucket boundaries ($b^k_0$, $b^k_1$, ..., $b^k_{30}$). The number of buckets is set to 32, matching the number of threads in a warp, allowing for efficient thread-level parallelism so that each thread processes a different bucket. \circled{2} Next, elements are gathered, starting from bucket 0, until the total count reaches $k_{chunk}$. \circled{3} If the number of elements in the current bucket exceeds the remaining spots for $k_{chunk}$, as in the case of bucket 9 in Figure~\ref{fig:topk}(b), random selection is used to fill the remaining spots. This random selection adds an additional layer of approximation but significantly reduces latency by avoiding exact sorting.

Determining proper bucket boundaries is crucial to minimize the approximation error introduced by random selection (\circled{3} in Figure~\ref{fig:topk}(b)). \name profiles the distribution of activation values using a small calibration set and aims to set boundary values that balance Top-K accuracy with the ability to handle a broader range of values. Placing finer-grained buckets around the expected $k$-th largest value generally improves accuracy, but limits the system's ability to handle out-of-distribution values. To address this, \name uses an offline analysis to determine two key boundaries, $b^k_0$ and $b^k_{15}$, from which the other boundaries are inferred (as shown in Figure~\ref{fig:boundary}). Let the distribution of activation values from the calibration set be $\mathbf{X} \in R^{N \times d_{in}}$, where $N$ is the size of the set. The boundary $b^k_{15}$ is set to the maximum of the $k$-th largest value across all vectors in $abs(\mathbf{X})$. The interval between 0 and $b^k_{15}$ is uniformly divided into 16 buckets ($b^{k}_{15}$, $b^{k}_{16}$, …, $b^{k}_{30}$), focusing on the range where the $k$-th largest value is most likely to occur. To handle out-of-distribution cases which can cause significant degradation in selection precision, \name assigns an additional 16 buckets for values beyond $b^k_{15}$. Specifically, $b^k_0$ is set to the maximum value in all $abs(\mathbf{X})$, and the range between $b^k_0$ and $b^k_{15}$ is also uniformly divided to form the remaining 16 buckets.




\begin{figure}[t]
    \centering
    \includegraphics[width=\columnwidth]{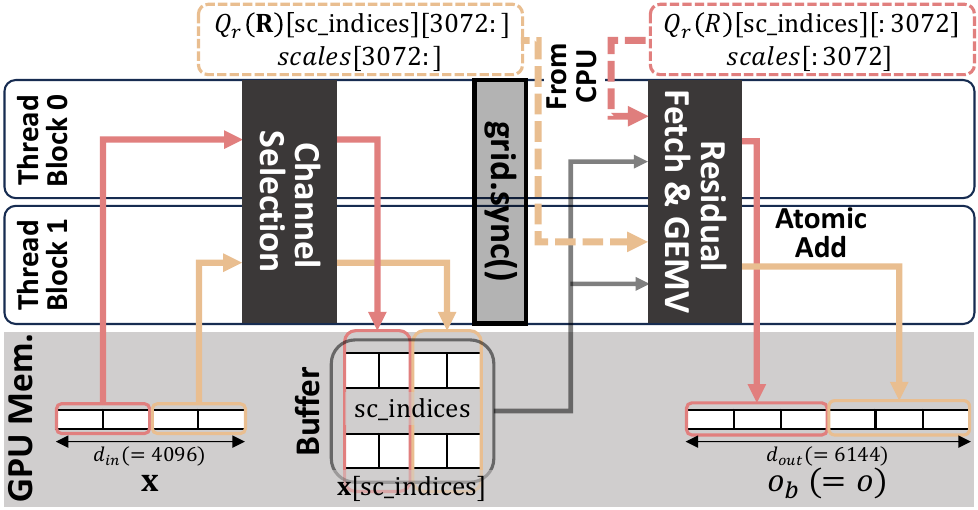}
    \caption{Fused kernel for dynamic error compensation.}
    \label{fig:fused_kernel}
\end{figure}

\para{Kernel Fusion}
\name extensively fuses all dynamic error compensation operations into a single kernel. 
Figure~\ref{fig:fused_kernel} visualizes the execution flow of the fused kernel, using an example where two thread blocks process a weight matrix of size 4096$\times$6144. In this example, only one channel is selected per each of the four chunks (i.e., $k_{chunk}=1$, $k=4$). Initially, each thread block sequentially processes two chunks, selecting a total of two channels (Step \circled{1} in Figure~\ref{fig:overview}(b)). The indices for the selected channels ($\texttt{sc\_indices}$) as well as the corresponding activation values ($\mathbf{x}[\texttt{sc\_indices}]$) are stored in GPU memory. Thread blocks are then synchronized using the grid-wide synchronization feature of the cooperative group~\cite{cooperative-groups}. This synchronization is required because each thread block processes a segment of all selected channels—not just a disjoint subset—when fetching quantized residuals from the CPU and performing GEMV (Steps~\circled{2} and \circled{3} in Figure~\ref{fig:overview}(b)). For example, thread block 0 processes $Q_{r}(\mathbf{R})[\texttt{sc\_indices}][:3072]$ as opposed to $Q_{r}(\mathbf{R})[\texttt{sc\_indices}[:2]][:]$. The thread block-level synchronization allows all thread blocks to have access to the complete set of $\texttt{sc\_indices}$ and $\mathbf{x}[\texttt{sc\_indices}]$. This partitioning scheme allows for efficient reduction in the residual GEMV without requiring extensive global synchronization. The results from the residual GEMV are directly added to the result of the base GEMV ($\mathbf{o_{b}}$) using atomic primitives, yielding the final output ($\mathbf{o}$) (Step \circled{4} in Figure~\ref{fig:overview}(b)).

\para{GPU Memory Overhead}
The buffer for $\texttt{sc\_indices}$ and $\mathbf{x}[\texttt{sc\_indices}]$ in the fused kernel is the only additional GPU memory usage of \name. Zero-copy from the CPU does not consume GPU memory. The bucket boundary values are not stored on the GPU; instead, only $b^k_0$ and $b^k_{15}$ are passed to the kernel as arguments. A single buffer can be reused for all linear layers if it is sized in accordance with the largest $k$. In the extreme case of fetching 10\% of the channels across all layers in Llama-3-8B, the maximum $k$ would be 1433, for the down projection layer. This calls for an 8.6 KB buffer ($1433\times(4+2)$), which is less than 0.0003\% of the model size, assuming 3-bit precision---essentially a negligible overhead.


\begin{figure*}[t]
    \centering
    \includegraphics[width=\textwidth]{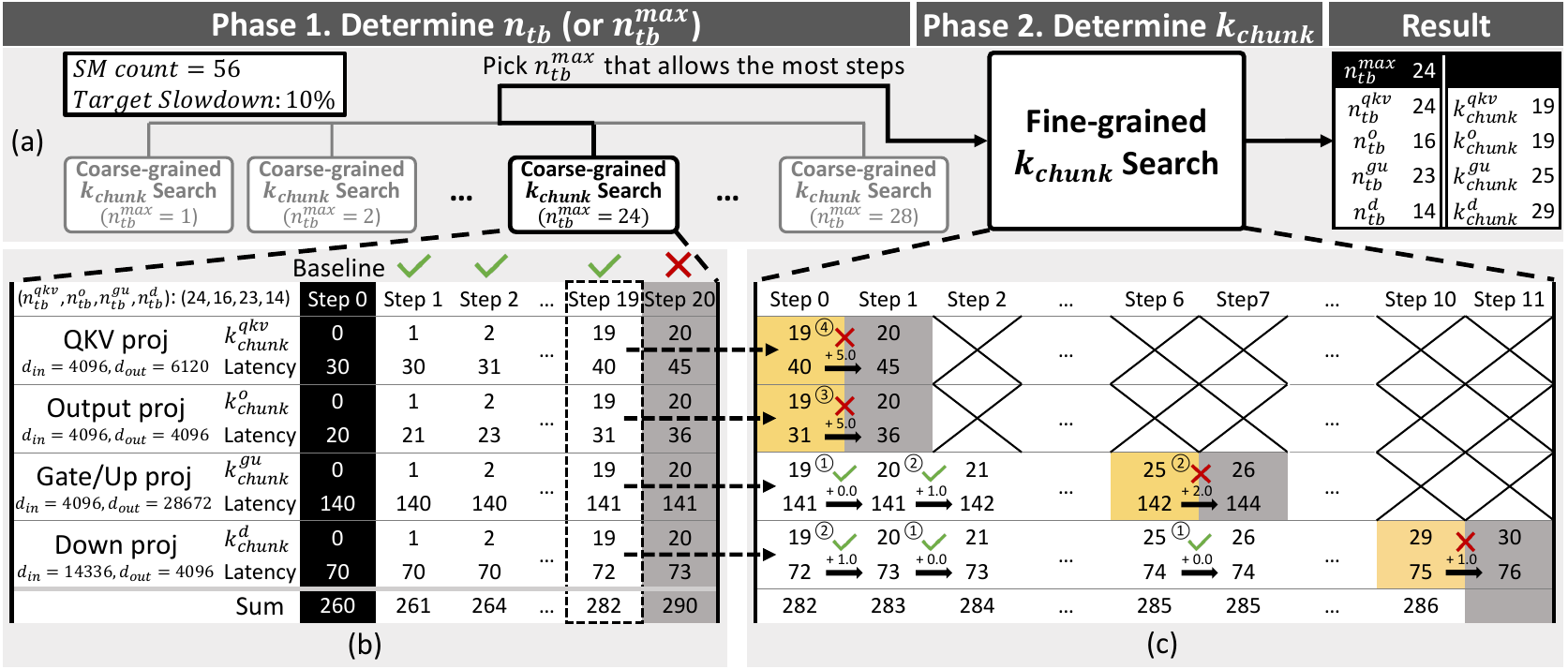}
    \caption{Parameter tuning process for \name, assuming a total of 56 SMs and a target slowdown rate of 10\%.}
    \label{fig:tuner}
\end{figure*}

\subsection{Parameter Tuner}
\label{sec:tuner}

\para{Necessity of Parameter Tuner}
Effective use of \name\ requires careful tuning two key parameters:

\begin{itemize}
  \item \textbf{$n_{tb}$}: Specifies the number of thread blocks used for dynamic error compensation.
  Since \name\ runs in parallel with the base GEMV, allocating too many thread blocks to dynamic error compensation can slow down the base GEMV computation,
  while allocating too few may underutilize PCIe bandwidth, as zero-copy transfers require GPU cores to issue memory requests.

  \item \textbf{$k_{chunk}$}: Specifies how many channels are compensated per chunk (1024 channels). 
  A larger $k_{chunk}$ improves model quality but may increase inference latency.
\end{itemize}


Choosing $n_{tb}$ and $k_{chunk}$ is challenging due to the large design space. The $n_{tb}$ value for each type of linear layer ($n^{qkv}_{tb}$, $n^{o}_{tb}$, $n^{gu}_{tb}$, $n^{d}_{tb}$) has multiple viable candidates, constrained by the dimensions of the corresponding weight matrices. For example, in Llama-3-8B, there are 9 possible candidates for $n^{qkv}_{tb}$ (1, 2, 3, 4, 5, 6, 8, 12, 24), while other layers also have multiple options. Selecting $k_{chunk}$ presents a greater challenge due to its broader range of possibilities. The $k_{chunk}$ value for each type of linear layer ($k^{qkv}_{chunk}$, $k^{o}_{chunk}$, $k^{gu}_{chunk}$, $k^{d}_{chunk}$) can be any integer less than a maximum determined by the available shared memory. Although platform-dependent, this upper bound is large (e.g., 367 with 48 KB per-block shared memory), leading to an expansive search space. This creates a combinatorial explosion of configuration options. Technical details on how to identify candidate values of $k_{chunk}$ and $n_{tb}$ for a given model and platform are presented at the end of this section.


To address this issue, we provide a \name\ tuner that suggests $n_{tb}$ and $k_{chunk}$ values based on a target slowdown rate. The tuner maximizes $k_{chunk}$ while keeping the total execution time---including base GEMV and dynamic error compensation across all linear layers---within the specified slowdown relative to the baseline (i.e., without compensation). This tuning is a one-time process for a given model-device pair.




\para{Parameter Tuning Process}
Figure~\ref{fig:tuner}(a) illustrates the tuning process, which consists of two phases: Phase 1 determines the $n_{tb}$ values, and Phase 2 determines the $k_{chunk}$ values.

In Phase 1, the tuner simplifies the search for $n_{tb}$ values for each layer by replacing it with the search for a single metaparameter, $n^{max}_{tb}$. $n^{max}_{tb}$ represents the upper limit on the number of thread blocks to use for dynamic error compensation. Each layer’s $n_{tb}$ is then set to the largest candidate below $n^{max}_{tb}$. Determining $n^{max}_{tb}$ involves testing values up to half of the total SM count, to reduce the search space. In the example shown in Figure~\ref{fig:tuner}, the GPU has 56 SMs, so values up to 28 are tested. For each tested $n^{max}_{tb}$, a coarse-grained $k_{chunk}$ search checks how many uniform increments to $k_{chunk}$ can be applied across all layers without exceeding the target slowdown rate. Figure~\ref{fig:tuner}(b) shows an example of a coarse-grained $k_{chunk}$ search for $n^{max}_{tb}=24$ (i.e., $(n^{qkv}_{tb},n^{o}_{tb},n^{gu}_{tb},n^{d}_{tb})=(24,16,23,14)$), yielding 19 valid steps. If no steps can be made for any $n^{max}_{tb}$ value, the tuner fixes $k_{chunk}$ to 0 for the layer with the smallest weight matrix and repeats the process, as smaller matrices are often most sensitive to increases in $k_{chunk}$.


After Phase 1, the tuner selects the $n^{max}_{tb}$ with the most steps and proceeds to a fine-grained $k_{chunk}$ search in Phase 2. Figure~\ref{fig:tuner}(c) shows the fine-grained $k$ search for the selected $n^{max}_{tb} = 24$. In this phase, not all $k_{chunk}$ values may increase together. At each step, the tuner increments $k_{chunk}$ for as many layers as possible, prioritizing those with smaller increases in execution time. For example, in Step 1, $k^{gu}_{chunk}$ is incremented first, followed by $k^{d}_{chunk}$. Step 1 stops at this point, as further increases to $k^{qkv}_{chunk}$ and $k^{o}_{chunk}$ would exceed the target; thus their final values are set (i.e., $k^{qkv}_{chunk}, k^{o}_{chunk}=19$). This process repeats until no further increments can be made for any layer.

\para{Technical Details: \texorpdfstring{$n_{tb}$}{ntb} and \texorpdfstring{$k_{chunk}$}{kchunk} Candidates}
The \(n_{tb}\) values considered during parameter tuning are those that have a meaningful impact on at least one of the two parts of the kernel execution: the approximate Top-K selection and the residual fetching. 
In the approximate Top-K selection part, the minimum processing granularity per thread block is one chunk. Therefore, increasing \(n_{tb}\) beyond the number of chunks does not have additional effect on performance. Thus, the set of \(n_{tb}\) values relevant to this part is:
\[
A = \left\{\,n \;\middle|\; 1 \le n \le \frac{d_{in}}{1024}\,\right\}.
\]
For residual fetching part, 4-bit residuals are transferred over PCIe in coalesced segments of 256 values (128 bytes), resulting in a total of \(s = d_{out} / 256\) segments. These segments are distributed across \(n_{tb}\) thread blocks, with each block processing \(\lceil s/n_{tb} \rceil\) segments. If multiple \(n_{tb}\) values result in the same number of segments per block (\(\lceil s/n_{tb} \rceil\)), only the smallest such value is considered and the rest are redundant. Excluding these cases, the candidate set relevant to this part is:
\[
B = \left\{\,n \;\middle|\; 1 \le n \le s,\;
\left\lceil \frac{s}{\left\lceil s / n \right\rceil} \right\rceil = n\,\right\}.
\]
The final candidate set for \(n_{tb}\) is the union of the two (i.e., \(N = A \cup B\)). 

Meanwhile, \(k_{chunk}\) is bounded by the shared memory limit. During the approximate Top-K selection part of the kernel, shared memory usage increases with the value of \(k_{chunk}\). Specifically, the shared memory usage for this part is:
\[
128 + 128 \times k_{chunk} + 2 \times 1024 \; \text{bytes}.
\]
Here, 128 bytes are used for integer counters that track the number of elements that fall into each of the 32 buckets; \(128 \times k_{chunk}\) bytes are used to temporarily store the indices of elements assigned to each bucket; and \(2 \times 1024\) bytes account for the input activation values in the chunk. The total must remain below the per-block shared memory limit (e.g., 49,152 bytes), which constrains the maximum allowable $k_{chunk}$ value.



%% file: contents/5-evaluation.tex
\section{Evaluation}
\label{sec:evaluation}


\begin{figure*}[t]
    \centering
    \includegraphics[width=\linewidth]{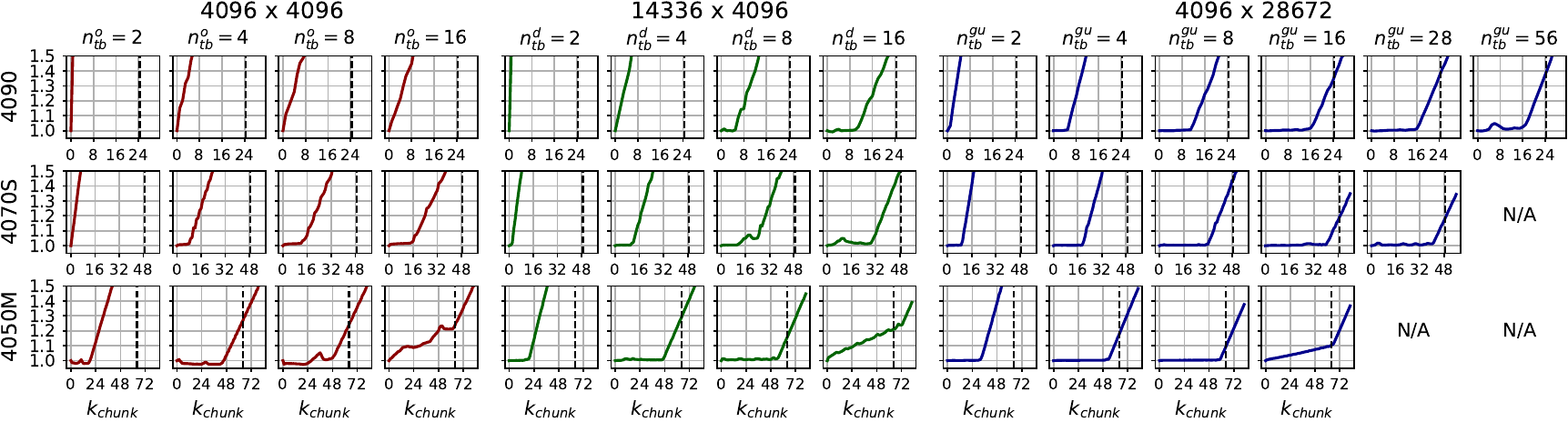}
    \caption{\rev{Execution time of base GEMV + \name with varying $k_{chunk}$ and $n_{tb}$, normalized to base GEMV execution time.}}
    \label{fig:profiler}
\end{figure*}

\subsection{GPU Kernel Benchmarks}
\label{sec:kernel_results}

\begin{table}[]
\centering
\resizebox{\linewidth}{!}{%

\begin{tabular}{lccccc}
\hline
\multicolumn{1}{l}{\textbf{GPU Name}} & \textbf{Memory Size} & \textbf{Memory BW} & \textbf{\# SM} & \textbf{PCIe BW} & \rev{$\boldsymbol{R_{bw}}$} \\ \hline
\multicolumn{6}{c}{\textbf{Desktop}} \\ \hline
\textbf{RTX 4090}     & 24 GB         & 1,008 GB/s         & 128               & 32 GB/s    &  \rev{32}  \\
\textbf{RTX 4080S}    & 16 GB         & 736 GB/s           & 80                & 32 GB/s    &  \rev{23}  \\
\textbf{RTX 4070S}    & 12 GB         & 504 GB/s           & 56                & 32 GB/s    &  \rev{16}  \\ \cline{1-6}
\multicolumn{6}{c}{\textbf{Laptop}} \\ \hline
\textbf{RTX 4070M}    & 8 GB          & 256 GB/s           & 36                & 16 GB/s    &  \rev{16}  \\
\textbf{RTX 4050M}    & 6 GB          & 192 GB/s           & 20                & 16 GB/s    &  \rev{12}  \\ \hline
\end{tabular}%
}
\caption{\rev{GPU specifications.}}
\label{tab:gpu_specs}
\end{table}

\para{Methodology}
In this section, we evaluate the \name GPU kernel on three consumer-grade GPUs: two desktop GPUs, RTX 4070 Super (RTX 4070S) and RTX 4090, and one laptop GPU, RTX 4050 Mobile (RTX 4050M), with their specifications detailed in Table~\ref{tab:gpu_specs}. \rev{Here, \rbw denotes the ratio of memory bandwidth to CPU-to-GPU (PCIe) bandwidth.} The evaluation considers GEMV operations of output, gate/up and down projection layers in Llama-3-8B-Instruct~\cite{llama-3}, assuming 3-bit bitwidth. For the base GEMV, we use the LUTGEMM kernel, ~\cite{lutgemm}, a state-of-the-art GEMV kernel for uniform quantization. Kernel times are measured using NVIDIA Nsight Systems.


\para{Expected Behavior}
The execution time of the \name kernel is expected to follow a piecewise linear function of $k_{chunk}$ with two distinct segments. In the first segment, corresponindg to small $k_{chunk}$ values, the latency remains nearly constant at the base GEMV time, as the compensation operations are fully hidden under the GEMV execution. In the second segment, once $k_{chunk}$ exceeds a certain threshold---the knee point---the execution time linearly increases with $k_{chunk}$, inidcating that the compensation operations are no longer fully hidden. \rev{Theoretically, the knee point is expected to occur when $k_{chunk}=1024 \times \nicefrac{1}{\text{\rbw}} \times \nicefrac{3}{4}$, representing the maximum amount of data transfer that can overlap with the base GEMV. Here, the base GEMV time is approximated by dividing the weight matrix size by the GPU memory bandwidth, and PCIe bandwidth is assumed to be fully utilized. The factor $\nicefrac{3}{4}$ accounts for the 3-bit quantization; for other bitwidths, this factor should be adjusted accordingly (e.g., $\nicefrac{4}{4} = 1$ for 4-bit quantization).}

\para{Results}
Figure~\ref{fig:profiler} shows the execution time of \name kernel (base GEMV + dynamic error compensation), normalized to the standalone execution time of the base GEMV, across varying $n_{tb}$ and $k_{chunk}$. \rev{Each subfigure also includes a vertical dotted line to mark the theoretical knee point derived from the analytical model (i.e., $1024\times\nicefrac{1}{\text{\rbw}} \times \nicefrac{3}{4}$).} All cases exhibit the expected two-segment piecewise linear behavior when $n_{tb}$ is properly set, except for the $4096 \times 4096$ case on RTX 4090, the fastest GPU evaluated. In this case, the base GEMV execution time is so short that even a small $k_{chunk}$ incurs overhead. 

From the other cases, three key observations can be made. \rev{First, a lower \rbw---indicating lower memory bandwidth and higher PCIe bandwidth---shifts the knee point to the right. Thus, RTX 4050M, with the lowest \rbw, supports the largest $k_{chunk}$ before the knee point, while RTX 4090, with the highest ratio, supports the smallest.} This trend is consistent with the ordering predicted by theoretical knee points. Second, the knee point is highly sensitive to $n_{tb}$, highlighting the importance of careful tuning. Generally, higher $n_{tb}$ values such as 8 or 16 delay the knee point. In contrast, small $n_{tb}$ values (e.g., 2) cause the knee point to appear too early or disappear, leading to suboptimal performance. However, increasing $n_{tb}$ can sometimes worsen results by slowing down the base GEMV, as discussed in Section~\ref{sec:tuner}. This effect is particularly noticeable on GPUs with fewer SMs, such as the 4050M. For instance, on the 4050M, $n_{tb}=8$ yields the best results, whereas increasing $n_{tb}$ to 16 leads to worse performance. Third, larger weight sizes enable higher $k_{chunk}$ with minimal latency overhead, due to increased time slack. \rev{With sufficiently large weight sizes (e.g., 4096$\times$28672) and a properly tuned $n_{tb}$, the actual knee point approaches the theoretical value. For example, on an RTX 4050M with a 4096$\times$28672 matrix and $n_{tb}=8$, the observed knee point is around $k_{chunk}=60$, compared to a theoretical prediction of 64.}

\subsection{Impact on Model Quality}
\label{sec:dec_results}
\para{Methodology}
We demonstrate \name’s quality improvements on 3-bit, 3.5-bit, and 4-bit versions of two instruction-tuned LLMs: Llama-3-8B-Instruct~\cite{llama-3} (hereafter referre to as Llama-3) and Phi-3-medium-4k-instruct (14B model, hereafter referred to as Phi-3)~\cite{phi-3}. For the 3.5-bit version, we adopt a block-wise bitwidth allocation, applying 3-bit quantization to half of the decoder blocks and 4-bit quantization to the remaining blocks. This follows a KL divergence-based sensitivity metric from prior work~\cite{zeroq}. 

As base quantization methods, we choose two state-of-the-art LLM quantization methods: AWQ~\cite{awq} and SqueezeLLM~\cite{sqllm}. AWQ is a \emph{uniform} quantization method that mitigates quantization errors by applying per-channel scaling to protect salient channels, which are identified through an offline analysis on a calibration dataset. SqueezeLLM employs a clustering-based \emph{non-uniform} quantization method that considers the sensitivity of each weight. 

We evaluate \name-augmented models across varying $k_{chunk}$, the number of channels compensated per chunk (1024 channels). We uniformly set $k_{chunk}$ to 8, 16, 32, 64 and 128 for all layers. \rev{Based on the results in Section~\ref{sec:kernel_results}, values up to 64 fall within the practical range—where latency overhead from dynamic error compensation may remain low depending on the platforms—while $k_{\text{chunk}}=128$ is included for completeness in evaluating upper-bound behavior.}

\para{Benchmarks} Following previous literature~\cite{quip, sqllm, awq, spqr}, we use perplexity on WikiText~\cite{wikitext} as the primary metric, as it reliably reflects quantized LLM quality~\cite{awq, scaling}. Additionally, we use BIG-Bench Hard (BBH)~\cite{bbh}, a collection of 23 challenging tasks in BIG-Bench~\cite{big-bench}, to assess the models' capabilities in problem solving. Chain-of-Thought (CoT) is enabled for all BBH evaluations~\cite{cot}. Lastly, we evaluate multi-turn conversation using MT-Bench~\cite{mt-bench}, where a strong LLM judges 80 responses, scoring each from 0 to 10. We use GPT-4o as the judge and report the average of three runs.

\begin{figure}[t]
    \centering
    \includegraphics[width=\linewidth]{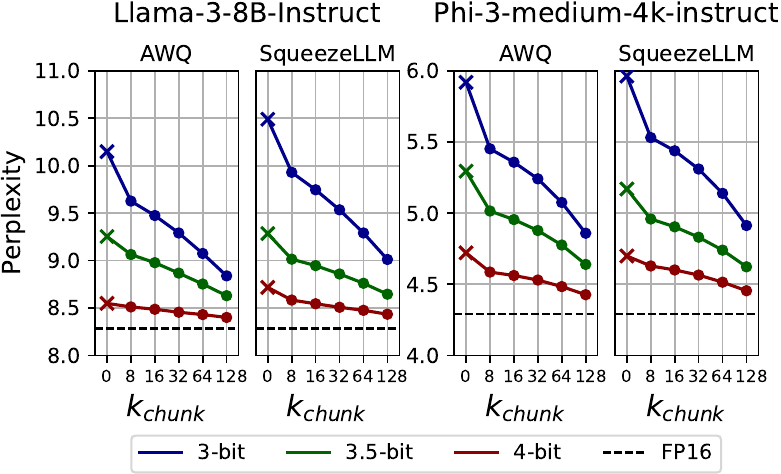}
    \caption{Perplexity on WikiText. The \texttt{x} markers correspond to baselines without \name ($k_{chunk} = 0$). Lower is better.}
    \label{fig:wikitext}
\end{figure}

\begin{figure}[t]
    \centering
    \includegraphics[width=\linewidth]{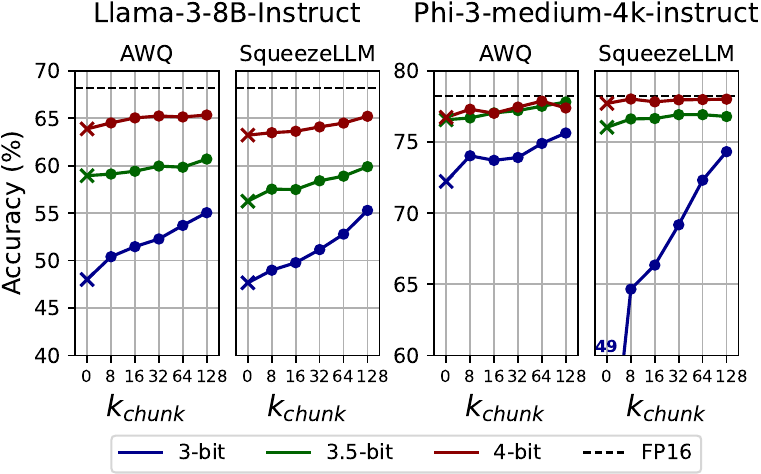}
    \caption{Accuracy on BBH. The \texttt{x} markers correspond to baselines without \name ($k_{chunk} = 0$). Higher is better.}
    \label{fig:bbh}
\end{figure}

\begin{figure}[t]
    \centering
    \includegraphics[width=\linewidth]{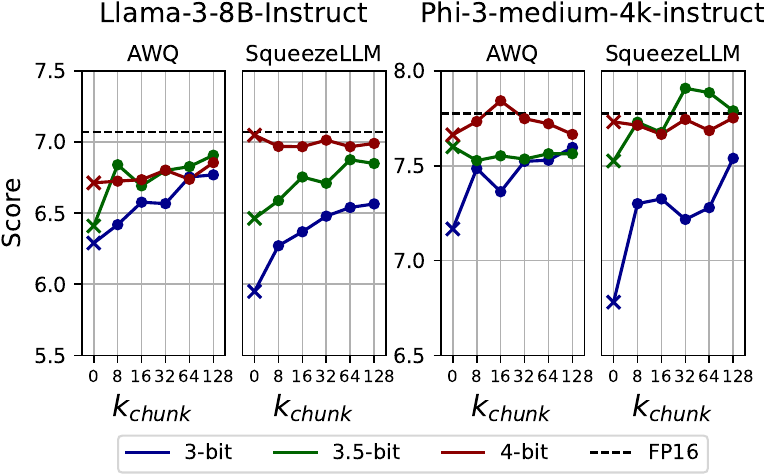}
    \caption{MT-Bench scores. The \texttt{x} markers correspond to baselines without \name ($k_{chunk} = 0$). Higher is better.}
    \label{fig:mtbench}
\end{figure}

\para{Perplexity on WikiText}
Figure~\ref{fig:wikitext} shows the perplexity results. Across all cases, a clear trend is observed: perplexity consistently decreases (i.e., model quality improves) as $k_{chunk}$ increases. In particular, for 3-bit models, significant improvements occur even at $k_{chunk}=8$. AWQ’s perplexity decreases from 10.15 to 9.63 for Llama-3 and from 5.96 to 5.53 for Phi-3, while SqueezeLLM’s perplexity drops from 10.49 to 9.93 for Llama-3 and from 5.92 to 5.45 for Phi-3. Meanwhile, the impact on 4-bit models is relatively less pronounced. This is expected, as 4-bit models are already close to full-precision, leaving less room for improvement. The 3.5-bit models follow an intermediate trend.

\para{BBH and MT-Bench} 
Results on BBH (Figure~\ref{fig:bbh}) follow the same trends as the perplexity results. The MT-Bench results (Figure~\ref{fig:mtbench}), on the other hand, demonstrate some different patterns. In cases where vanilla quantized models without \name ($k=0$) already achieve scores very close to those of the FP16 model---such as all 4-bit cases and the AWQ 3.5-bit model of Phi-3---the scores remain unchanged, oscillating around the baseline score. For the remaining cases, \name significantly improves scores even with a small $k_{chunk}$ (e.g., 8), like in other benchmarks; further increases in $k_{chunk}$, however, do not always yield noticeable improvements. These patterns may be attributed to the coarse-grained rubric of this benchmark, which assigns integer scores ranging from 0 to 10 for each task. This coarse-grained scoring may miss subtle improvements when the potential gain is small.



\begin{figure}[t]
    \centering
    \includegraphics[width=\linewidth]{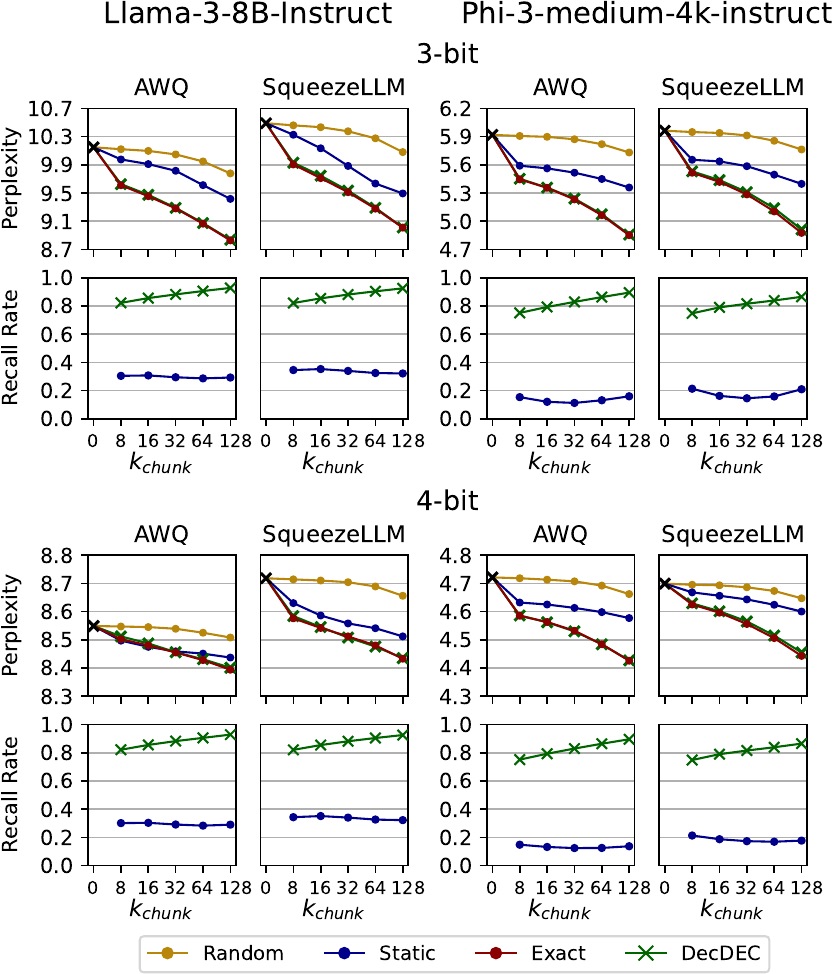}
    \caption{\rev{Comparison against random, static, and exact channel selection. Perplexity is shown on the top rows (lower is better), and recall is shown on the bottom rows (higher is better). The \texttt{x} markers in black correspond to baselines without \name{} ($k_{chunk} = 0$).}}
    \label{fig:vsstatictopk}
\end{figure}



\para{\rev{Effectiveness of \name's Channel Selection}}
Figure~\ref{fig:vsstatictopk} compares \name\ with three variants using different channel selection mechanisms. To assess the benefits of dynamic selection, we include \emph{Random} (blue line), which selects channels randomly, and \emph{Static} (yellow line), which statically select channels by Hessian-based ranking with a calibration set (using exact sorting) following prior work~\cite{owq}. To isolate the effect of \name's Top-K approximation, we include \emph{Exact} (red line), which uses true Top-K channels. Alongside perplexity results, we also report the average recall rate of Static and \name relative to Exact.

\rev{While Static improves over Random, it underperforms \name. \name\ achieves lower perplexity than Static while using 4× or even 8× fewer channels (e.g., $k_{\text{chunk}} = 32$ or $16$ vs.\ $128$), highlighting the effectiveness of incorporating input-awareness. Meanwhile, the perplexity gap between \name and Exact is minimal, with almost overlapping curves. These results are explained by the recall rates: while \name\ achieves around 80\% recall relative to Exact, the recall of Static falls significantly short, at around 30\% or below.}

\begin{table}[t]
\centering
\fontsize{32}{40}\selectfont
\resizebox{\linewidth}{!}{%
\begin{tabular}{|c|cccc|cccc|cccc|cccc|}
\multicolumn{1}{c}{} & \multicolumn{8}{c}{\fontsize{40}{50}\selectfont\textbf{Llama-3-8B-Instruct}} & \multicolumn{8}{c}{\fontsize{40}{50}\selectfont\textbf{Phi-3-medium-4k-instruct}} \\
\cline{2-17}
\multicolumn{1}{c|}{} & \multicolumn{4}{c|}{\textbf{AWQ 3-bit}} & \multicolumn{4}{c|}{\textbf{SqueezeLLM 3-bit}} & \multicolumn{4}{c|}{\textbf{AWQ 3-bit}} & \multicolumn{4}{c|}{\textbf{SqueezeLLM 3-bit}} \\
\hline
\textbf{$k_{chunk}$} & \textbf{2-bit} & \textbf{4-bit} & \textbf{8-bit} & \textbf{FP16} & \textbf{2-bit} & \textbf{4-bit} & \textbf{8-bit} & \textbf{FP16} & \textbf{2-bit} & \textbf{4-bit} & \textbf{8-bit} & \textbf{FP16} & \textbf{2-bit} & \textbf{4-bit} & \textbf{8-bit} & \textbf{FP16} \\
\hline
2 & \cellcolor{gray!40} & \cellcolor{gray!40} & \cellcolor{gray!40} & \cellcolor{red!30}9.83 & \cellcolor{gray!40} & \cellcolor{gray!40} & \cellcolor{gray!40} & \cellcolor{red!30}10.10 & \cellcolor{gray!40} & \cellcolor{gray!40} & \cellcolor{gray!40} & \cellcolor{red!30}5.56 & \cellcolor{gray!40} & \cellcolor{gray!40} & \cellcolor{gray!40} & \cellcolor{red!30}5.63 \\
4 & \cellcolor{gray!40} & \cellcolor{gray!40} & \cellcolor{red!30}9.72 & \cellcolor{orange!30}9.72 & \cellcolor{gray!40} & \cellcolor{gray!40} & \cellcolor{red!30}9.97 & \cellcolor{orange!30}9.97 & \cellcolor{gray!40} & \cellcolor{gray!40} & \cellcolor{red!30}5.51 & \cellcolor{orange!30}5.51 & \cellcolor{gray!40} & \cellcolor{gray!40} & \cellcolor{red!30}5.55 & \cellcolor{orange!30}5.56 \\
8 & \cellcolor{gray!40} & \cellcolor{red!30}\textbf{9.63} & \cellcolor{orange!30}9.58 & \cellcolor{yellow!30}9.58 & \cellcolor{gray!40} & \cellcolor{red!30}\textbf{9.93} & \cellcolor{orange!30}9.79 & \cellcolor{yellow!30}9.78 & \cellcolor{gray!40} & \cellcolor{red!30}\textbf{5.45} & \cellcolor{orange!30}5.45 & \cellcolor{yellow!30}5.45 & \cellcolor{gray!40} & \cellcolor{red!30}5.53 & \cellcolor{orange!30}5.47 & \cellcolor{yellow!30}5.47 \\
16 & \cellcolor{red!30}\textbf{9.63} & \cellcolor{orange!30}\textbf{9.47} & \cellcolor{yellow!30}9.43 & \cellcolor{green!30}9.43 & \cellcolor{red!30}9.95 & \cellcolor{orange!30}\textbf{9.75} & \cellcolor{yellow!30}9.59 & \cellcolor{green!30}9.58 & \cellcolor{red!30}5.47 & \cellcolor{orange!30}\textbf{5.36} & \cellcolor{yellow!30}5.35 & \cellcolor{green!30}5.35 & \cellcolor{red!30}\textbf{5.51} & \cellcolor{orange!30}5.44 & \cellcolor{yellow!30}5.36 & \cellcolor{green!30}5.37 \\
32 & \cellcolor{orange!30}9.49 & \cellcolor{yellow!30}\textbf{9.29} & \cellcolor{green!30}9.25 & \cellcolor{cyan!30}9.25 & \cellcolor{orange!30}9.80 & \cellcolor{yellow!30}\textbf{9.54} & \cellcolor{green!30}9.36 & \cellcolor{cyan!30}9.36 & \cellcolor{orange!30}\textbf{5.36} & \cellcolor{yellow!30}5.24 & \cellcolor{green!30}5.23 & \cellcolor{cyan!30}5.23 & \cellcolor{orange!30}\textbf{5.41} & \cellcolor{yellow!30}5.31 & \cellcolor{green!30}5.23 & \cellcolor{cyan!30}5.24 \\
64 & \cellcolor{yellow!30}9.31 & \cellcolor{green!30}\textbf{9.07} & \cellcolor{cyan!30}9.03 & \cellcolor{gray!40} & \cellcolor{yellow!30}9.59 & \cellcolor{green!30}\textbf{9.29} & \cellcolor{cyan!30}9.11 & \cellcolor{gray!40} & \cellcolor{yellow!30}\textbf{5.22} & \cellcolor{green!30}5.08 & \cellcolor{cyan!30}5.06 & \cellcolor{gray!40} & \cellcolor{yellow!30}\textbf{5.28} & \cellcolor{green!30}5.14 & \cellcolor{cyan!30}5.06 & \cellcolor{gray!40} \\
128 & \cellcolor{green!30}9.12 & \cellcolor{cyan!30}\textbf{8.84} & \cellcolor{gray!40} & \cellcolor{gray!40} & \cellcolor{green!30}9.36 & \cellcolor{cyan!30}\textbf{9.01} & \cellcolor{gray!40} & \cellcolor{gray!40} & \cellcolor{green!30}\textbf{5.04} & \cellcolor{cyan!30}4.86 & \cellcolor{gray!40} & \cellcolor{gray!40} & \cellcolor{green!30}\textbf{5.10} & \cellcolor{cyan!30}\textbf{4.91} & \cellcolor{gray!40} & \cellcolor{gray!40} \\
256 & \cellcolor{cyan!30}8.94 & \cellcolor{gray!40} & \cellcolor{gray!40} & \cellcolor{gray!40} & \cellcolor{cyan!30}9.14 & \cellcolor{gray!40} & \cellcolor{gray!40} & \cellcolor{gray!40} & \cellcolor{cyan!30}\textbf{4.83} & \cellcolor{gray!40} & \cellcolor{gray!40} & \cellcolor{gray!40} & \cellcolor{cyan!30}\textbf{4.91} & \cellcolor{gray!40} & \cellcolor{gray!40} & \cellcolor{gray!40} \\
\hline
\end{tabular}
}
\caption{Impact of residual bitwidth. Lower is better.}
\label{tab:bitwidth}
\end{table}

\para{Impact of Residual Bitwidth}
Table~\ref{tab:bitwidth} presents the impact of residual bitwidth selection. In addition to the default 4-bit setting, we evaluate 2-bit, 8-bit, and full-precision (FP16) residuals with varying $k_{chunk}$ for 3-bit models by perplexity on WikiText. Cells with the same color indicate cases where the total data transfer via PCIe is approximately equivalent. For example, $k_{chunk}=8$ with 4-bit residuals requires a similar data transfer amount as the following combinations: $k_{chunk}=16$ at 2-bit, $k_{chunk}=4$ at 8-bit, and $k_{chunk}=2$ at FP16. The best results within each color group are highlighted. Across all cases, a residual bitwidth of 4 either achieves the best or comes very close to it, supporting our default setting.

\subsection{End-to-End Evaluation}
\label{sec:case_study}
\para{Methodology}
In this section, we present case studies demonstrating how \name advances model quality while minimizing latency increases across three desktop GPUs---RTX 4090, 4080 Super (4080S), and 4070 Super (4070S)---as well as two laptop GPUs---RTX 4070 Mobile (4070M), and 4050 Mobile (4050M). The specifications for these GPUs are listed in Table~\ref{tab:gpu_specs}. We run the tuner for \name with four target slowdown rates (2.5\%, 5\%, 10\%, 20\%), and evaluate perplexity on WikiText as well as end-to-end inference latency with the resulting configurations. For the base GEMV kernel, we use LUTGEMM~\cite{lutgemm} for AWQ and Any-Precision LLM~\cite{any-precision-llm} for SqueezeLLM—both state-of-the-art kernels for uniform and non-uniform quantization, respectively. We integrate \name into a PyTorch-based inference pipeline optimized with the torch.compile feature~\cite{torch-compile} and measure the average time taken per token generation over 1024 tokens. 

For the 3.5-bit models, we do not run the tuner separately. Instead, we construct the configuration by combining the tuning results with the 3-bit and 4-bit models. Specifically, we use the configuration from tuning with the 3-bit model for layers quantized to 3 bits, and the configuration from tuning with the 4-bit model for layers quantized to 4 bits---both assuming the same target slowdown rate.

\begin{figure*}[t]
    \centering
    \includegraphics[width=0.9\textwidth]{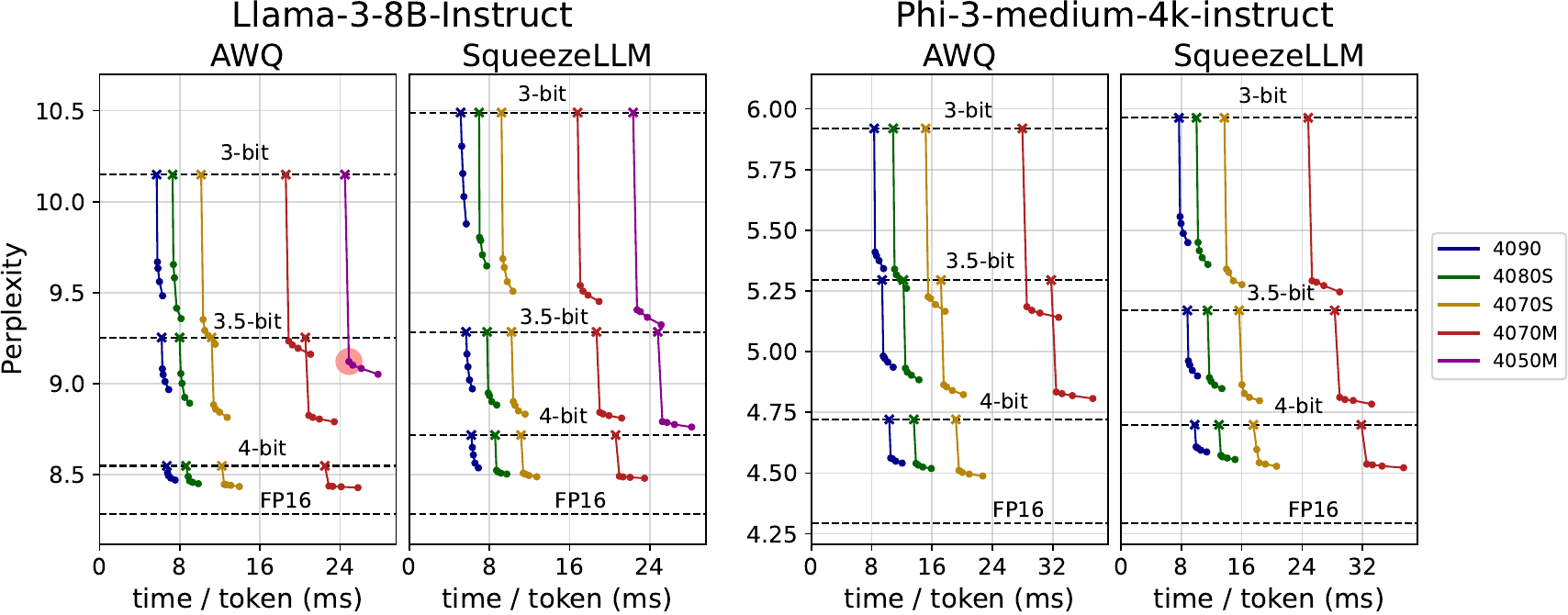}
    \caption{Perplexity against time per token on various NVIDIA GPUs. The \texttt{x} markers indicate baseline values, where \name is not applied ($k_{chunk} = 0$). Subsequent markers show DecDEC results on target slowdown rates 2.5\%, 5\%, 10\%, and 20\%.}
    \label{fig:tradeoff}
\end{figure*}

\begin{table*}[h!]
\centering
\fontsize{12}{16}\selectfont
\resizebox{\linewidth}{!}{
\begin{tabular}{|cc|cc|cc|cc|cc|}
\cline{3-10}
\multicolumn{2}{c|}{} &\multicolumn{4}{c|}{\textbf{Llama-3-8B-Instruct}} &\multicolumn{4}{c|}{\textbf{Phi-3-medium-4k-instruct}} \\
\cline{3-10}
\multicolumn{2}{c|}{} &\multicolumn{2}{c|}{\textbf{AWQ 3-bit}} &\multicolumn{2}{c|}{\textbf{SqueezeLLM 3-bit}} &\multicolumn{2}{c|}{\textbf{AWQ 3-bit}} &\multicolumn{2}{c|}{\textbf{SqueezeLLM 3-bit}} \\
\hline
\textbf{GPU} & \textbf{Target} & \textbf{Tuner Results} & \textbf{Slowdown} & \textbf{Tuner Results} & \textbf{Slowdown} & \textbf{Tuner Results} & \textbf{Slowdown} & \textbf{Tuner Results} & \textbf{Slowdown} \\
\hline
\multirow{4}{*}{\textbf{4090}} & \textbf{2.5\%} & 24 / (4, 4, 8, 9) & 1.3\% & 56 / (2, 0, 2, 2) & 2.0\% & 20 / (8, 6, 13, 13) & 1.5\% & 47 / (5, 2, 7, 10) & 1.8\% \\
 & \textbf{5\%} & 24 / (5, 7, 9, 10) & 2.2\% & 56 / (1, 1, 2, 6) & 4.0\% & 47 / (10, 11, 13, 13) & 3.8\% & 35 / (6, 6, 9, 10) & 3.5\% \\
 & \textbf{10\%} & 24 / (10, 11, 11, 11) & 4.9\% & 38 / (5, 4, 5, 7) & 6.1\% & 35 / (15, 15, 15, 14) & 7.8\% & 47 / (10, 11, 13, 11) & 7.3\% \\
 & \textbf{20\%} & 24 / (15, 15, 16, 15) & 10.4\% & 28 / (10, 10, 10, 10) & 10.8\% & 20 / (18, 19, 18, 18) & 15.0\% & 35 / (16, 16, 15, 14) & 15.3\% \\
\hline
\multirow{4}{*}{\textbf{4080S}} & \textbf{2.5\%} & 6 / (9, 5, 9, 6) & 1.1\% & 24 / (10, 11, 14, 16) & 0.6\% & 30 / (18, 17, 18, 20) & 1.7\% & 30 / (14, 14, 15, 16) & 2.0\% \\
 & \textbf{5\%} & 16 / (10, 10, 11, 9) & 2.5\% & 24 / (14, 14, 14, 14) & 2.2\% & 30 / (21, 21, 23, 21) & 3.9\% & 30 / (17, 17, 20, 19) & 3.7\% \\
 & \textbf{10\%} & 24 / (18, 20, 23, 18) & 5.5\% & 24 / (18, 19, 18, 17) & 4.7\% & 35 / (24, 25, 24, 24) & 8.3\% & 35 / (22, 22, 23, 21) & 7.6\% \\
 & \textbf{20\%} & 24 / (26, 27, 25, 24) & 11.5\% & 38 / (23, 24, 22, 23) & 10.8\% & 35 / (30, 32, 28, 29) & 16.2\% & 35 / (27, 27, 25, 25) & 15.1\% \\
\hline
\multirow{4}{*}{\textbf{4070S}} & \textbf{2.5\%} & 28 / (25, 26, 26, 26) & 2.0\% & 19 / (16, 15, 24, 21) & 1.5\% & 20 / (31, 31, 37, 32) & 2.1\% & 20 / (25, 26, 30, 28) & 2.1\% \\
 & \textbf{5\%} & 24 / (31, 31, 35, 29) & 3.7\% & 24 / (21, 22, 24, 24) & 3.0\% & 20 / (35, 35, 37, 34) & 4.2\% & 20 / (30, 30, 30, 29) & 4.2\% \\
 & \textbf{10\%} & 24 / (35, 35, 36, 34) & 7.1\% & 24 / (28, 30, 30, 27) & 6.2\% & 24 / (39, 40, 41, 38) & 8.5\% & 24 / (34, 35, 35, 33) & 8.6\% \\
 & \textbf{20\%} & 28 / (44, 44, 41, 42) & 13.9\% & 24 / (35, 36, 34, 35) & 12.4\% & 20 / (45, 46, 44, 45) & 17.0\% & 20 / (40, 41, 38, 39) & 16.9\% \\
\hline
\multirow{4}{*}{\textbf{4070M}} & \textbf{2.5\%} & 16 / (38, 39, 39, 40) & 1.7\% & 12 / (28, 28, 35, 29) & 1.6\% & 18 / (41, 41, 41, 42) & 2.1\% & 17 / (33, 35, 35, 34) & 2.1\% \\
 & \textbf{5\%} & 16 / (42, 43, 42, 42) & 3.4\% & 16 / (33, 34, 36, 33) & 3.1\% & 18 / (44, 46, 43, 44) & 4.4\% & 18 / (37, 38, 37, 37) & 4.3\% \\
 & \textbf{10\%} & 16 / (46, 46, 44, 45) & 6.6\% & 16 / (37, 38, 37, 37) & 6.1\% & 18 / (46, 48, 46, 46) & 8.3\% & 18 / (40, 41, 39, 39) & 8.5\% \\
 & \textbf{20\%} & 16 / (50, 50, 49, 50) & 13.4\% & 16 / (42, 42, 41, 41) & 12.7\% & 18 / (53, 53, 50, 51) & 17.2\% & 18 / (44, 44, 43, 43) & 16.7\% \\
\hline
\multirow{4}{*}{\textbf{4050M}} & \textbf{2.5\%} & 8 / (55, 56, 58, 55) & 1.7\% & 7 / (45, 45, 50, 44) & 1.7\% & OOM & - & OOM & - \\
 & \textbf{5\%} & 8 / (59, 59, 59, 58) & 3.3\% & 9 / (48, 48, 48, 48) & 3.3\% & OOM & - & OOM & - \\
 & \textbf{10\%} & 10 / (62, 62, 62, 62) & 6.6\% & 10 / (52, 53, 52, 52) & 6.3\% & OOM & - & OOM & - \\
 & \textbf{20\%} & 10 / (70, 70, 68, 68) & 13.5\% & 10 / (59, 60, 57, 58) & 12.6\% & OOM & - & OOM & - \\
\hline
\end{tabular}
}
\caption{Tuner results (\text{\scriptsize{$n^{max}_{tb} / (k^{qkv}_{chunk}, k^{o}_{chunk}, k^{gu}_{chunk}, k^{d}_{chunk})$}}) for four target slowdown rates \text{(2.5\%, 5\%, 10\%, 20\%)} and corresponding actual slowdown rates for 3-bit Llama-3 and Phi-3. Phi-3 is out of memory (OOM) on the 4050M GPU.}
\label{tab:autotuner}
\end{table*}

\para{Results}
Table~\ref{tab:autotuner} lists the configurations obtained from the tuner along with the actual end-to-end slowdowns compared to the baseline. We report only the results for 3-bit as similar trends are observed for the 3.5-bit and 4-bit configurations. In all cases, the actual slowdown is below the target rate. This is expected as the tuner configures parameters conservatively. The tuner targets only the kernel times of linear operations, while other operations outside the linear layers (e.g., attention, normalization) also contribute to overall inference time. Consistent with the observations in Section~\ref{sec:kernel_results}, in general, selected $k$ values are higher for GPUs with greater PCIe-to-GPU memory bandwidth ratio (4050M $>$ 4070M $\simeq$ 4070S $>$ 4080S $>$ 4090). 


Figure~\ref{fig:tradeoff} shows the trends of end-to-end inference latency versus perplexity. All Phi-3 cases, as well as AWQ 3.5-bit, 4-bit, and SqueezeLLM 4-bit cases of Llama-3 face out-of-memory issues on the 4050M, and are thus excluded. Similarly, the AWQ 4-bit case of Phi-3 for 4070M is also excluded. In each line in Figure~\ref{fig:tradeoff}, the \texttt{x} marker represents the baseline, while subsequent markers show the results for \name on the four target slowdown rates, in increasing order. 

For all cases, \name demonstrates promising Pareto-optimal trade-offs between model quality and inference latency. The results for target slowdown rates of 2.5\% and 5\% are particularly impressive, while 10\%–20\% show diminishing returns. On platforms with high PCIe-to-GPU memory bandwidth ratios, such as the 4070S, 4070M, and 4050M, \name at 2.5\% slowdown on 3-bit models sometimes outperforms 3.5-bit baselines. In these cases, \name yields \emph{Pareto-dominant} solutions by excelling in model quality, latency, and memory. Examples include AWQ Llama-3 (4070M, 4050M) and AWQ Phi-3 (4070S, 4070M). A particularly noteworthy case is the AWQ 3-bit Llama-3 on 4050M, where \name reduces perplexity from 10.15 to 9.12 with only a 1.7\% latency slowdown (highlighted by a red circle). This outperforms the 3.5-bit baseline, which is infeasible on 4050M without \name due to memory limits. This case highlights \name's effectiveness in pushing the boundaries of quantized LLMs within memory capacity constraints.

\begin{figure}[t]
    \centering
    \includegraphics[width=\linewidth]{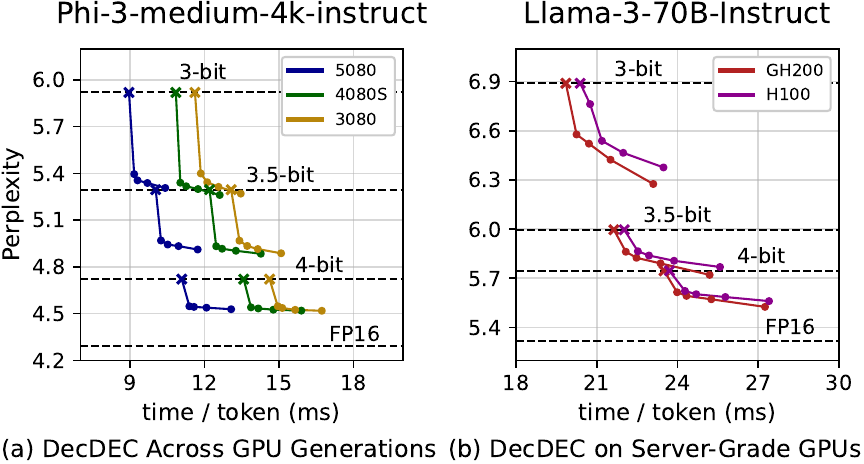}
    \caption{\rev{Perplexity vs. time per token: (a) across different GPU generations and (b) on server-grade GPUs.}}
    \label{fig:tradeoff_server_and_gen}
\end{figure}

\subsection{\rev{Evaluation across GPU Generations}}
\label{sec:gpu_generation}
\rev{We evaluate the robustness of \name across GPU generations, as both GPU memory bandwidth and PCIe bandwidth, two key factors influencing its effectiveness, may increase with newer architectures. To this end, we use three consumer-grade GPUs from the same product class across different generations---RTX 3080, 4080S, and 5080---whose specifications are listed in Table~\ref{tab:gpu_generation}. The \rbw values (lower is better) remain essentially unchanged from the 3080 to the 4080S and even decrease from the 4080S to the 5080. Figure~\ref{fig:tradeoff_server_and_gen}(a) plots the perplexity of AWQ-quantized Phi-3 with \name\ augmentation against end-to-end latency on these GPUs, using the methodology described in Section~\ref{sec:case_study}. The improvements delivered by \name are comparable across all three cards, confirming its robustness across generations.}

\subsection{\rev{Applicability to Server-Grade GPUs}}
\label{sec:server}
\rev{Although \name primarily targets single-batch inference on client (edge) devices, this section evaluates its effectiveness on server-grade GPUs. We measure perplexity versus end-to-end latency using an AWQ-quantized Llama-3-70B-Instruct model augmented with \name on the H100 SMX5 and GH200, following the methodology described in Section~\ref{sec:case_study}. Both GPUs provide 3.36 TB/s of memory bandwidth; however, the GH200’s 450 GB/s NVLink-C2C far exceeds the H100’s 64 GB/s PCIe, yielding a much lower \rbw.}

\rev{Figure~\ref{fig:tradeoff_server_and_gen}(b) shows the results. \name improves perplexity on both GPUs with minimal latency overhead. However, the GH200’s advantage is smaller than the \rbw gap suggests. \name assumes a DRAM-bound GEMV, where reallocating SMs for error compensation do not impact GEMV latency. On server GPUs, however, quantized GEMV (e.g., LUTGEMM~\cite{lutgemm}) is L1-bound, not DRAM-bound. Since L1 throughput scales with active SMs, reallocating SMs increases GEMV latency, limiting \name's benefit despite low \rbw. Enhancing quantized GEMV kernels for server-grade GPUs by mitigating L1 bottlenecks could unlock further gains.}

\begin{table}[]
\centering
\resizebox{\linewidth}{!}{%
\begin{tabular}{lccc}
          & \textbf{Memory Bandwidth} & \textbf{PCIe Bandwidth} & $\boldsymbol{R_{bw}}$ \\ \hline
\textbf{RTX 5080}  & 960 GB/s     & 64 GB/s               & 15                                            \\
\textbf{RTX 4080S} & 736 GB/s     & 32 GB/s                & 23                                             \\
\textbf{RTX 3080}  & 760 GB/s     & 32 GB/s               & 24 \\ \hline                                  
\end{tabular}%
}
\caption{\rev{80-class GPU specifications across generations.}}
\label{tab:gpu_generation}
\end{table}


%% file: contents/6-related-works.tex
\section{Related Work}
\label{sec:related_works}
\para{GPU Implementations for Weight-Only Quantization}
Numerous studies have proposed efficient GPU implementations tailored for weight-only quantization of LLMs. 
LUTGEMM~\cite{lutgemm} replaces the expensive dequantization with simple LUT operations. Marlin~\cite{marlin} supports FP16-INT4 GEMM across various batch sizes, and Quant-LLM~\cite{fp6} introduces a GPU kernel that efficiently handles non-power-of-2 bitwidths (e.g., 6-bit). FLUTE~\cite{flute} provides a specialized kernel for non-uniform quantization. Any-Precision LLM~\cite{any-precision-llm} suggests a memory-efficient kernel for adaptive bitwidth selection.  These implementations can seamlessly integrate with \name, benefiting from its dynamic error compensation.

\para{LLM Inference with External Memory} 
Several works address GPU memory limitations in LLM inference by leveraging external memories like CPU memory or disk. 
DeepSpeed-Inference~\cite{deepspeed} and FlexGen~\cite{flexgen} focus on throughput-oriented, out-of-core LLM inference. Pre-gated MoE~\cite{pregated} retrieves the parameters of only the activated experts in MoE models from CPU memory. LLM-in-a-Flash~\cite{llm-in-a-flash} introduces a flash memory-based inference system exploiting activation sparsity. InfiniGen~\cite{infinigen} offloads the key-value cache to CPU memory. PowerInfer~\cite{powerinfer} proposes a GPU-CPU hybrid inference engine that leverages activation sparsity in ReLU-based models. Though these approaches share the same goal as \name in aiming to extend GPU memory, they address distinct scenarios with different challenges and opportunities. 

\para{LLM Compression Methods}
In addition to quantization, pruning can improve the efficiency of LLM inference~\cite{pruning-1, pruning-2, pruning-3}. Other techniques, including knowledge distillation~\cite{distillation-1, distillation-2} and low-rank decomposition~\cite{asvd,decomposition-2}, offer additional ways to compress LLMs. These methods are orthogonal to quantization and, consequently, to \name.

%% file: contents/7-conclusion.tex
\section{Conclusion}
\label{sec:conclusion}
We propose \name, an inference scheme for low-bit LLMs that improves model quality by correcting quantization errors through selective retrieval of residuals stored in CPU memory. By focusing on dynamically identified salient channels at each decoding step, \name maximizes error compensation within a limited transfer volume. \name significantly improves quality of quantized LLMs with minimal overheads.

%% file: main.bbl
\begin{thebibliography}{10}

\bibitem{h100_datasheet}
{NVIDIA H100 Tensor Core GPU}.
\newblock \url{https://resources.nvidia.com/en-us-data-center-overview-mc/en-us-data-center-overview/nvidia-tensor-core-gpu-datasheet}, 2024.

\bibitem{bucket}
Tolu Alabi, Jeffrey~D. Blanchard, Bradley Gordon, and Russel Steinbach.
\newblock Fast \textit{k}-selection algorithms for graphics processing units.
\newblock {\em ACM J. Exp. Algorithmics}, 2012.

\bibitem{llm-in-a-flash}
Keivan Alizadeh, Seyed~Iman Mirzadeh, Dmitry Belenko, S.~Khatamifard, Minsik Cho, Carlo~C Del~Mundo, Mohammad Rastegari, and Mehrdad Farajtabar.
\newblock {LLM} in a flash: Efficient large language model inference with limited memory.
\newblock In {\em Proceedings of the 62nd Annual Meeting of the Association for Computational Linguistics}, 2024.

\bibitem{deepspeed}
Reza~Yazdani Aminabadi, Samyam Rajbhandari, Ammar~Ahmad Awan, Cheng Li, Du~Li, Elton Zheng, Olatunji Ruwase, Shaden Smith, Minjia Zhang, Jeff Rasley, and Yuxiong He.
\newblock Deepspeed-inference: enabling efficient inference of transformer models at unprecedented scale.
\newblock In {\em Proceedings of the International Conference on High Performance Computing, Networking, Storage and Analysis}, 2022.

\bibitem{torch-compile}
Jason Ansel, Edward Yang, Horace He, Natalia Gimelshein, Animesh Jain, Michael Voznesensky, Bin Bao, Peter Bell, David Berard, Evgeni Burovski, Geeta Chauhan, Anjali Chourdia, Will Constable, Alban Desmaison, Zachary DeVito, Elias Ellison, Will Feng, Jiong Gong, Michael Gschwind, Brian Hirsh, Sherlock Huang, Kshiteej Kalambarkar, Laurent Kirsch, Michael Lazos, Mario Lezcano, Yanbo Liang, Jason Liang, Yinghai Lu, C.~K. Luk, Bert Maher, Yunjie Pan, Christian Puhrsch, Matthias Reso, Mark Saroufim, Marcos~Yukio Siraichi, Helen Suk, Shunting Zhang, Michael Suo, Phil Tillet, Xu~Zhao, Eikan Wang, Keren Zhou, Richard Zou, Xiaodong Wang, Ajit Mathews, William Wen, Gregory Chanan, Peng Wu, and Soumith Chintala.
\newblock Pytorch 2: Faster machine learning through dynamic python bytecode transformation and graph compilation.
\newblock In {\em Proceedings of the 29th ACM International Conference on Architectural Support for Programming Languages and Operating Systems}, 2024.

\bibitem{quantizable}
Yelysei Bondarenko, Markus Nagel, and Tijmen Blankevoort.
\newblock Quantizable transformers: Removing outliers by helping attention heads do nothing.
\newblock In {\em Advances in Neural Information Processing Systems}, 2023.

\bibitem{fewshot}
Tom~B. Brown, Benjamin Mann, Nick Ryder, Melanie Subbiah, Jared Kaplan, Prafulla Dhariwal, Arvind Neelakantan, Pranav Shyam, Girish Sastry, Amanda Askell, Sandhini Agarwal, Ariel Herbert-Voss, Gretchen Krueger, Tom Henighan, Rewon Child, Aditya Ramesh, Daniel~M. Ziegler, Jeffrey Wu, Clemens Winter, Christopher Hesse, Mark Chen, Eric Sigler, Mateusz Litwin, Scott Gray, Benjamin Chess, Jack Clark, Christopher Berner, Sam McCandlish, Alec Radford, Ilya Sutskever, and Dario Amodei.
\newblock Language models are few-shot learners.
\newblock In {\em Advances in Neural Information Processing Systems}, volume~33, 2020.

\bibitem{zeroq}
Yaohui Cai, Zhewei Yao, Zhen Dong, Amir Gholami, Michael~W Mahoney, and Kurt Keutzer.
\newblock Zeroq: A novel zero shot quantization framework.
\newblock In {\em Proceedings of the IEEE/CVF Conference on Computer Vision and Pattern Recognition}, 2020.

\bibitem{quip}
Jerry Chee, Yaohui Cai, Volodymyr Kuleshov, and Christopher~De Sa.
\newblock Qu{IP}: 2-bit quantization of large language models with guarantees.
\newblock In {\em Advances in Neural Information Processing Systems}, 2023.

\bibitem{llmint8}
Tim Dettmers, Mike Lewis, Younes Belkada, and Luke Zettlemoyer.
\newblock Llm.int8(): 8-bit matrix multiplication for transformers at scale.
\newblock In {\em Proceedings of the 36th International Conference on Neural Information Processing Systems}, 2024.

\bibitem{qlora}
Tim Dettmers, Artidoro Pagnoni, Ari Holtzman, and Luke Zettlemoyer.
\newblock {QL}o{RA}: Efficient finetuning of quantized {LLM}s.
\newblock In {\em Thirty-seventh Conference on Neural Information Processing Systems}, 2023.

\bibitem{spqr}
Tim Dettmers, Ruslan~A. Svirschevski, Vage Egiazarian, Denis Kuznedelev, Elias Frantar, Saleh Ashkboos, Alexander Borzunov, Torsten Hoefler, and Dan Alistarh.
\newblock Sp{QR}: A sparse-quantized representation for near-lossless {LLM} weight compression.
\newblock In {\em The Twelfth International Conference on Learning Representations}, 2024.

\bibitem{scaling}
Tim Dettmers and Luke Zettlemoyer.
\newblock The case for 4-bit precision: k-bit inference scaling laws.
\newblock In {\em Proceedings of the 40th International Conference on Machine Learning}, 2023.

\bibitem{layerwise-2}
Zhen Dong, Zhewei Yao, Daiyaan Arfeen, Amir Gholami, Michael~W Mahoney, and Kurt Keutzer.
\newblock Hawq-v2: Hessian aware trace-weighted quantization of neural networks.
\newblock In {\em Advances in Neural Information Processing Systems}, 2020.

\bibitem{layerwise-1}
Zhen Dong, Zhewei Yao, Amir Gholami, Michael Mahoney, and Kurt Keutzer.
\newblock Hawq: Hessian aware quantization of neural networks with mixed-precision.
\newblock In {\em 2019 IEEE/CVF International Conference on Computer Vision}, 2019.

\bibitem{llama-3}
Abhimanyu Dubey, Abhinav Jauhri, Abhinav Pandey, Abhishek Kadian, Ahmad Al-Dahle, Aiesha Letman, Akhil Mathur, Alan Schelten, Amy Yang, and et~al Angela~Fan.
\newblock The llama 3 herd of models, 2024.

\bibitem{aqlm}
Vage Egiazarian, Andrei Panferov, Denis Kuznedelev, Elias Frantar, Artem Babenko, and Dan Alistarh.
\newblock Extreme compression of large language models via additive quantization.
\newblock In {\em Proceedings of the 41st International Conference on Machine Learning}, 2024.

\bibitem{pruning-1}
Elias Frantar and Dan Alistarh.
\newblock Sparsegpt: massive language models can be accurately pruned in one-shot.
\newblock In {\em Proceedings of the 40th International Conference on Machine Learning}, 2023.

\bibitem{gptq}
Elias Frantar, Saleh Ashkboos, Torsten Hoefler, and Dan Alistarh.
\newblock {OPTQ}: Accurate quantization for generative pre-trained transformers.
\newblock In {\em The Eleventh International Conference on Learning Representations}, 2023.

\bibitem{marlin}
Elias Frantar, Roberto~L. Castro, Jiale Chen, Torsten Hoefler, and Dan Alistarh.
\newblock Marlin: Mixed-precision auto-regressive parallel inference on large language models.
\newblock In {\em Proceedings of the 30th ACM SIGPLAN Annual Symposium on Principles and Practice of Parallel Programming}, 2025.

\bibitem{pile}
Leo Gao, Stella Biderman, Sid Black, Laurence Golding, Travis Hoppe, Charles Foster, Jason Phang, Horace He, Anish Thite, Noa Nabeshima, Shawn Presser, and Connor Leahy.
\newblock The pile: An 800gb dataset of diverse text for language modeling, 2020.

\bibitem{distillation-2}
Yuxian Gu, Li~Dong, Furu Wei, and Minlie Huang.
\newblock Mini{LLM}: Knowledge distillation of large language models.
\newblock In {\em The Twelfth International Conference on Learning Representations}, 2024.

\bibitem{flute}
Han Guo, William Brandon, Radostin Cholakov, Jonathan Ragan-Kelley, Eric~P. Xing, and Yoon Kim.
\newblock Fast matrix multiplications for lookup table-quantized {LLM}s.
\newblock In {\em Findings of the Association for Computational Linguistics: EMNLP 2024}, 2024.

\bibitem{rethinking}
Jung~Hwan Heo, Jeonghoon Kim, Beomseok Kwon, Byeongwook Kim, Se~Jung Kwon, and Dongsoo Lee.
\newblock Rethinking channel dimensions to isolate outliers for low-bit weight quantization of large language models.
\newblock In {\em The Twelfth International Conference on Learning Representations}, 2024.

\bibitem{distillation-1}
Cheng-Yu Hsieh, Chun-Liang Li, Chih-kuan Yeh, Hootan Nakhost, Yasuhisa Fujii, Alex Ratner, Ranjay Krishna, Chen-Yu Lee, and Tomas Pfister.
\newblock Distilling step-by-step! outperforming larger language models with less training data and smaller model sizes.
\newblock In {\em Findings of the Association for Computational Linguistics: ACL 2023}, 2023.

\bibitem{slim-llm}
Wei Huang, Haotong Qin, Yangdong Liu, Yawei Li, Xianglong Liu, Luca Benini, Michele Magno, and Xiaojuan Qi.
\newblock Slim-llm: Salience-driven mixed-precision quantization for large language models.
\newblock 2024.

\bibitem{llama-3-quant}
Wei Huang, Xingyu Zheng, Xudong Ma, Haotong Qin, Chengtao Lv, Hong Chen, Jie Luo, Xiaojuan Qi, Xianglong Liu, and Michele Magno.
\newblock An empirical study of llama3 quantization: From llms to mllms, 2024.

\bibitem{pregated}
Ranggi Hwang, Jianyu Wei, Shijie Cao, Changho Hwang, Xiaohu Tang, Ting Cao, and Mao Yang.
\newblock Pre-gated moe: An algorithm-system co-design for fast and scalable mixture-of-expert inference.
\newblock In {\em 2024 ACM/IEEE 51st Annual International Symposium on Computer Architecture}, 2024.

\bibitem{peqa}
Jeonghoon Kim, Jung~Hyun Lee, Sungdong Kim, Joonsuk Park, Kang~Min Yoo, Se~Jung Kwon, and Dongsoo Lee.
\newblock Memory-efficient fine-tuning of compressed large language models via sub-4-bit integer quantization.
\newblock In {\em Advances in Neural Information Processing Systems}, 2023.

\bibitem{sqllm}
Sehoon Kim, Coleman Hooper, Amir Gholami, Zhen Dong, Xiuyu Li, Sheng Shen, Michael~W Mahoney, and Kurt Keutzer.
\newblock Squeeze{LLM}: Dense-and-sparse quantization.
\newblock In {\em Proceedings of the 41st International Conference on Machine Learning}, 2024.

\bibitem{vllm}
Woosuk Kwon, Zhuohan Li, Siyuan Zhuang, Ying Sheng, Lianmin Zheng, Cody~Hao Yu, Joseph Gonzalez, Hao Zhang, and Ion Stoica.
\newblock Efficient memory management for large language model serving with pagedattention.
\newblock In {\em Proceedings of the 29th Symposium on Operating Systems Principles}, 2023.

\bibitem{owq}
Changhun Lee, Jungyu Jin, Taesu Kim, Hyungjun Kim, and Eunhyeok Park.
\newblock Owq: Outlier-aware weight quantization for efficient fine-tuning and inference of large language models.
\newblock In {\em Proceedings of the AAAI Conference on Artificial Intelligence}, 2024.

\bibitem{emnlp}
Janghwan Lee, Minsoo Kim, Seungcheol Baek, Seok Hwang, Wonyong Sung, and Jungwook Choi.
\newblock Enhancing computation efficiency in large language models through weight and activation quantization.
\newblock In {\em Proceedings of the 2023 Conference on Empirical Methods in Natural Language Processing}, 2023.

\bibitem{infinigen}
Wonbeom Lee, Jungi Lee, Junghwan Seo, and Jaewoong Sim.
\newblock {InfiniGen}: Efficient generative inference of large language models with dynamic {KV} cache management.
\newblock In {\em 18th USENIX Symposium on Operating Systems Design and Implementation}, 2024.

\bibitem{awq}
Ji~Lin, Jiaming Tang, Haotian Tang, Shang Yang, Wei-Ming Chen, Wei-Chen Wang, Guangxuan Xiao, Xingyu Dang, Chuang Gan, and Song Han.
\newblock Awq: Activation-aware weight quantization for on-device llm compression and acceleration.
\newblock In {\em Proceedings of Machine Learning and Systems}, 2024.

\bibitem{llmqat}
Zechun Liu, Barlas Oguz, Changsheng Zhao, Ernie Chang, Pierre Stock, Yashar Mehdad, Yangyang Shi, Raghuraman Krishnamoorthi, and Vikas Chandra.
\newblock {LLM}-{QAT}: Data-free quantization aware training for large language models.
\newblock In {\em Findings of the Association for Computational Linguistics: ACL 2024}, 2024.

\bibitem{pruning-3}
Xinyin Ma, Gongfan Fang, and Xinchao Wang.
\newblock Llm-pruner: On the structural pruning of large language models.
\newblock In {\em Advances in Neural Information Processing Systems}, 2023.

\bibitem{wikitext}
Stephen Merity, Caiming Xiong, James Bradbury, and Richard Socher.
\newblock Pointer sentinel mixture models, 2016.

\bibitem{phi-3}
Microsoft.
\newblock Phi-3 technical report: A highly capable language model locally on your phone, 2024.

\bibitem{pytorch_direct}
Seung~Won Min, Kun Wu, Sitao Huang, Mert Hidayeto\u{g}lu, Jinjun Xiong, Eiman Ebrahimi, Deming Chen, and Wen-mei Hwu.
\newblock Large graph convolutional network training with gpu-oriented data communication architecture.
\newblock In {\em Proc. VLDB Endow.}, 2021.

\bibitem{zero-copy}
NVIDIA.
\newblock {CUDA C++ Best Practices Guide}.
\newblock \url{https://docs.nvidia.com/cuda/cuda-c- best-practices-guide/index.html}, 2024.

\bibitem{cooperative-groups}
NVIDIA.
\newblock {CUDA C++ Programming Guide}.
\newblock \url{https://docs.nvidia.com/cuda/cuda-c-programming-guide/}, 2024.

\bibitem{gpt-4}
OpenAI.
\newblock Gpt-4 technical report, 2024.

\bibitem{lutgemm}
Gunho Park, Baeseong Park, Minsub Kim, Sungjae Lee, Jeonghoon Kim, Beomseok Kwon, Se~Jung Kwon, Byeongwook Kim, Youngjoo Lee, and Dongsoo Lee.
\newblock Lut-gemm: Quantized matrix multiplication based on luts for efficient inference in large-scale generative language models.
\newblock In {\em The Twelfth International Conference on Learning Representations}, 2024.

\bibitem{any-precision-llm}
Yeonhong Park, Jake Hyun, SangLyul Cho, Bonggeun Sim, and Jae~W. Lee.
\newblock Any-precision llm: Low-cost deployment of multiple, different-sized llms.
\newblock In {\em Proceedings of the 41st International Conference on Machine Learning}, 2024.

\bibitem{pearson}
Carl Pearson, Abdul Dakkak, Sarah Hashash, Cheng Li, I-Hsin Chung, Jinjun Xiong, and Wen-Mei Hwu.
\newblock Evaluating characteristics of cuda communication primitives on high-bandwidth interconnects.
\newblock In {\em Proceedings of the 2019 ACM/SPEC International Conference on Performance Engineering}, 2019.

\bibitem{c4}
Colin Raffel, Noam Shazeer, Adam Roberts, Katherine Lee, Sharan Narang, Michael Matena, Yanqi Zhou, Wei Li, and Peter~J. Liu.
\newblock Exploring the limits of transfer learning with a unified text-to-text transformer.
\newblock {\em J. Mach. Learn. Res.}, 2020.

\bibitem{decomposition-2}
Varun~Srivastava Rajarshi~Saha and Mert Pilanci.
\newblock {Matrix Compression via Randomized Low Rank and Low Precision Factorization}.
\newblock In {\em Advances in Neural Information Processing Systems}, 2023.

\bibitem{omniquant}
Wenqi Shao, Mengzhao Chen, Zhaoyang Zhang, Peng Xu, Lirui Zhao, Zhiqian Li, Kaipeng Zhang, Peng Gao, Yu~Qiao, and Ping Luo.
\newblock Omniquant: Omnidirectionally calibrated quantization for large language models.
\newblock 2023.

\bibitem{flexgen}
Ying Sheng, Lianmin Zheng, Binhang Yuan, Zhuohan Li, Max Ryabinin, Beidi Chen, Percy Liang, Christopher R\'{e}, Ion Stoica, and Ce~Zhang.
\newblock Flexgen: high-throughput generative inference of large language models with a single gpu.
\newblock In {\em Proceedings of the 40th International Conference on Machine Learning}, 2023.

\bibitem{powerinfer}
Yixin Song, Zeyu Mi, Haotong Xie, and Haibo Chen.
\newblock Powerinfer: Fast large language model serving with a consumer-grade gpu.
\newblock In {\em Proceedings of the ACM SIGOPS 30th Symposium on Operating Systems Principles}, 2024.

\bibitem{big-bench}
Aarohi Srivastava, Abhinav Rastogi, Abhishek Rao, Abu Awal~Md Shoeb, Abubakar Abid, Adam Fisch, Adam~R Brown, Adam Santoro, Aditya Gupta, Adri{\`a} Garriga-Alonso, et~al.
\newblock Beyond the imitation game: Quantifying and extrapolating the capabilities of language models.
\newblock 2022.

\bibitem{pruning-2}
Mingjie Sun, Zhuang Liu, Anna Bair, and J.~Zico Kolter.
\newblock A simple and effective pruning approach for large language models.
\newblock 2023.

\bibitem{bbh}
Mirac Suzgun, Nathan Scales, Nathanael Sch{\"a}rli, Sebastian Gehrmann, Yi~Tay, Hyung~Won Chung, Aakanksha Chowdhery, Quoc~V Le, Ed~H Chi, Denny Zhou, , and Jason Wei.
\newblock Challenging big-bench tasks and whether chain-of-thought can solve them.
\newblock 2022.

\bibitem{gemini}
Gemini Team.
\newblock Gemini 1.5: Unlocking multimodal understanding across millions of tokens of context, 2024.

\bibitem{llama}
Hugo Touvron, Thibaut Lavril, Gautier Izacard, Xavier Martinet, Marie-Anne Lachaux, Timothée Lacroix, Baptiste Rozière, Naman Goyal, Eric Hambro, Faisal Azhar, Aurelien Rodriguez, Armand Joulin, Edouard Grave, and Guillaume Lample.
\newblock {LLaMA}: Open and efficient foundation language models, 2023.

\bibitem{quip-sharp}
Albert Tseng, Jerry Chee, Qingyao Sun, Volodymyr Kuleshov, and Christopher~De Sa.
\newblock Qu{IP}\${\textbackslash}\#\$: Even better {LLM} quantization with hadamard incoherence and lattice codebooks.
\newblock In {\em Forty-first International Conference on Machine Learning}, 2024.

\bibitem{exllamav2}
turboderp.
\newblock Ex{L}lama{V}2.
\newblock \url{https://github.com/turboderp/exllamav2}.

\bibitem{attention}
Ashish Vaswani, Noam Shazeer, Niki Parmar, Jakob Uszkoreit, Llion Jones, Aidan~N. Gomez, \L{}ukasz Kaiser, and Illia Polosukhin.
\newblock Attention is all you need.
\newblock In {\em Proceedings of the 31st International Conference on Neural Information Processing Systems}, 2017.

\bibitem{efficient}
Zhongwei Wan, Xin Wang, Che Liu, Samiul Alam, Yu~Zheng, Jiachen Liu, Zhongnan Qu, Shen Yan, Yi~Zhu, Quanlu Zhang, Mosharaf Chowdhury, and Mi~Zhang.
\newblock Efficient large language models: A survey, 2024.

\bibitem{channelwise-1}
Zhe Wang, Jie Lin, Xue Geng, Mohamed M.~Sabry Aly, and Vijay Chandrasekhar.
\newblock Rdo-q: Extremely fine-grained channel-wise quantization via rate-distortion optimization.
\newblock In {\em Computer Vision -- ECCV 2022}, 2022.

\bibitem{cot}
Jason Wei, Xuezhi Wang, Dale Schuurmans, Maarten Bosma, Brian Ichter, Fei Xia, Ed~H. Chi, Quoc~V. Le, and Denny Zhou.
\newblock Chain-of-thought prompting elicits reasoning in large language models.
\newblock In {\em Proceedings of the 36th International Conference on Neural Information Processing Systems}, 2024.

\bibitem{fp6}
Haojun Xia, Zhen Zheng, Xiaoxia Wu, Shiyang Chen, Zhewei Yao, Stephen Youn, Arash Bakhtiari, Michael Wyatt, Donglin Zhuang, Zhongzhu Zhou, Olatunji Ruwase, Yuxiong He, and Shuaiwen~Leon Song.
\newblock {Quant-LLM}: Accelerating the serving of large language models via {FP6-Centric} {Algorithm-System} {Co-Design} on modern {GPUs}.
\newblock In {\em 2024 USENIX Annual Technical Conference}, 2024.

\bibitem{smoothquant}
Guangxuan Xiao, Ji~Lin, Mickael Seznec, Hao Wu, Julien Demouth, and Song Han.
\newblock {S}mooth{Q}uant: Accurate and efficient post-training quantization for large language models.
\newblock In {\em Proceedings of the 40th International Conference on Machine Learning}, 2023.

\bibitem{zeroquant}
Zhewei Yao, Reza Yazdani~Aminabadi, Minjia Zhang, Xiaoxia Wu, Conglong Li, and Yuxiong He.
\newblock Zeroquant: Efficient and affordable post-training quantization for large-scale transformers.
\newblock In {\em Advances in Neural Information Processing Systems}, 2022.

\bibitem{orca}
Gyeong-In Yu, Joo~Seong Jeong, Geon-Woo Kim, Soojeong Kim, and Byung-Gon Chun.
\newblock Orca: A distributed serving system for {Transformer-Based} generative models.
\newblock In {\em 16th USENIX Symposium on Operating Systems Design and Implementation}, 2022.

\bibitem{asvd}
Zhihang Yuan, Yuzhang Shang, Yue Song, Qiang Wu, Yan Yan, and Guangyu Sun.
\newblock Asvd: Activation-aware singular value decomposition for compressing large language models, 2023.

\bibitem{mt-bench}
Lianmin Zheng, Wei-Lin Chiang, Ying Sheng, Siyuan Zhuang, Zhanghao Wu, Yonghao Zhuang, Zi~Lin, Zhuohan Li, Dacheng Li, Eric~P. Xing, Hao Zhang, Joseph~E. Gonzalez, and Ion Stoica.
\newblock Judging llm-as-a-judge with mt-bench and chatbot arena.
\newblock In {\em Proceedings of the 37th International Conference on Neural Information Processing Systems}, 2024.

\bibitem{survey}
Xunyu Zhu, Jian Li, Yong Liu, Can Ma, and Weiping Wang.
\newblock A survey on model compression for large language models, 2024.

\end{thebibliography}
